\documentclass[10pt,journal,compsoc]{IEEEtran}
             
\ifCLASSOPTIONcompsoc
  \usepackage[nocompress]{cite}
\else
  \usepackage{cite}
\fi
\ifCLASSINFOpdf
\else
\fi
\usepackage{amsmath}
\usepackage{balance}
\usepackage{booktabs} % For formal tables
\usepackage{graphicx}
\usepackage{multirow}
\usepackage{pdfsync}
\usepackage{rotating}
\usepackage{soul}
\usepackage{url}
\usepackage[utf8]{inputenc}
\usepackage[table]{xcolor}
\usepackage{subcaption}
\usepackage[export]{adjustbox}

\usepackage{epstopdf-base}
\csname define@key\endcsname{Gin}{pdfpng}[]{%
\epstopdfDeclareGraphicsRule{.pdf}{png}{.png}{%
convert -density 100 ##1 \OutputFile
}%
}
\newif\ifdraft
\newcommand{\fabscomment}[1]{\ifdraft{\leavevmode\color{red}{[FS]: {#1}}}\else{\vspace{0ex}}\fi}

\newcommand{\bblue}[1]{\ifdraft{\leavevmode\color{black}{#1}}\else{\textcolor{black}{#1}}\fi}

\def\shadow{\cellcolor[HTML]{D6D6D6}}

\DeclareMathOperator{\tfidf}{TFIDF}
\DeclareMathOperator{\bm25}{BM25}
\DeclareMathOperator{\tw}{TW}
\DeclareMathOperator{\dd}{DD}
\DeclareMathOperator{\cd}{CD}
\DeclareMathOperator{\ig}{IG}
\DeclareMathOperator{\gr}{GR}
\DeclareMathOperator{\rf}{RF}
\DeclareMathOperator{\tpr}{TPR}
\DeclareMathOperator{\fpr}{FPR}

\DeclareMathOperator{\ReLU}{ReLU}

\usepackage{ifthen}

\newcounter{IssueCounter}

\begin{document}

\title{Learning to Weight for Text Classification}

\author{Alejandro Moreo Fern\'andez, Andrea Esuli, and~Fabrizio
Sebastiani \IEEEcompsocitemizethanks{\IEEEcompsocthanksitem All
authors are with Istituto di Scienza e Tecnologie dell'Informazione,
Consiglio Nazionale delle Ricerche,
56124 Pisa, Italy.\protect\\
E-mail: \{firstname,lastname\}@isti.cnr.it } \thanks{Manuscript
received \today.}}%; revised August 26, 2015.}}

\markboth{}%
{Shell \MakeLowercase{\textit{et al.}}: Bare Advanced Demo of
IEEEtran.cls for IEEE Computer Society Journals}
\IEEEtitleabstractindextext{%
\begin{abstract}
  In information retrieval (IR) and related tasks, term weighting
  approaches typically consider the frequency of the term in the
  document and in the collection in order to compute a score
  reflecting the importance of the term for the document.  In tasks
  characterized by the presence of training data (such as text
  classification) it seems logical that the term weighting function
  should take into account the distribution (as estimated from
  training data) of the term across the classes of interest. Although
  ``supervised term weighting'' approaches that use this intuition
  have been described before, they have failed to show consistent
  improvements. In this article we analyse the possible reasons for
  this failure, and call consolidated assumptions into question.
  Following this criticism we propose a novel supervised term
  weighting approach that, instead of relying on any predefined
  formula, learns a term weighting function optimised on the training
  set of interest; we dub this approach \emph{Learning to Weight}
  (LTW).
  % \fabscomment{Apprendere una funzione per ogni dataset potrebbe
  % essere considerato eccessivamente costoso; e.g., nel Learning to
  % Rank si apprende una funzione di ranking, ma non specifica di un
  % dataset. Potrebbe essere pensabile qualcosa del genere anche nel
  % nostro caso?}
  The experiments that we run on several well-known benchmarks, and
  using different learning methods, show that our method outperforms
  previous term weighting approaches in text classification.
  \fabscomment{Checked, Aug 14.}
\end{abstract}

}

\maketitle

\IEEEdisplaynontitleabstractindextext \IEEEpeerreviewmaketitle

% -------------------------------------------------

\ifCLASSOPTIONcompsoc
\IEEEraisesectionheading{\section{Introduction}\label{sec:introduction}}
\else
\section{Introduction}
\label{sec:introduction}
\fi \IEEEPARstart{I}{n} information retrieval (IR) and related
disciplines, where each textual document is represented as a vector of
terms (a.k.a.\ ``features''), term weighting consists of computing a
numerical score that reflects the importance of a given term $t$ for a
given document $d$ \cite{Salton88}.  Once the designer has decided
what should constitute a ``term'' (e.g., a word, or the morphological
root of a word, or a character $n$-gram, etc.), and has thus decided
what a vector consists of in the system, the term weighting method is
responsible for filling out the vector that will represent a specific
document. Once filled, this vector is fed to the module responsible
for computing document-document similarity (when \textit{ad hoc}
search, or text clustering, are the tasks of interest), or to the
module responsible for training a classifier, or to the classifier
itself (in the case of text classification). Different term weighting
methods generate different vectors for the same document, thus
attributing to the document different semantic
interpretations. ``Good'' term weighting methods are thus of
fundamental importance for delivering good
search/clustering/classification accuracy.

Term weighting approaches widely used in IR include TFIDF and BM25;
loosely speaking, both methods (as long as practically all other term
weighting methods in the literature) rely on the same principles,
i.e., (a) that terms that occur more frequently in the document (i.e.,
terms with high ``term frequency'') are more relevant to the document,
and (b) that terms that occur in fewer documents (i.e., terms with
high ``inverse document frequency'') are more relevant \textit{tout
court}. Different interpretations of these principles lead to the many
term weighting formulae proposed in the literature
\cite{Paltoglou:2010uq,Salton88,Zobel98}. However, practically all
such formulae share a common structure.  To see this, let us take one
of the many variants of TFIDF, i.e.,
\begin{equation}
  \tfidf(t,d,D)=\log(1+ f_{td})\cdot\log\frac{|D|}{n}
  \label{eq:logtfidf}
\end{equation}
\noindent and let us take BM25,
i.e., %\fabscomment{Ci sono davvero molte varianti di BM25?}
\begin{equation}
  \begin{aligned}
    \bm25(t,d,D) & = \frac{(k_1+1)\cdot
    f_{td}}{k_1\left((1-b)+b\cdot\frac{dl}{avdl}\right)+f_{td}} \\ &
    \hspace{5em} \cdot\log{\left(\frac{|D|-n+0.5}{n+0.5}\right)}
    \label{eq:bm25}
  \end{aligned}
\end{equation}
\noindent as examples. Here, $t$ is the term and $d$ is the document
for which the score is being computed, $D$ is the collection of
documents (or: the set of training documents, in a text classification
context), $f_{td}$ is the raw frequency (i.e., the number of
occurrences) of term $t$ in document $d$,
% $N$ is the number of documents,
$n\equiv |\{d' \in D : t \in d'\}|$ is the number of documents
containing the term, $dl$ and $avdl$ are the length of $d$ (i.e., the
number of term occurrences that $d$ contains) and the average length
of the documents in $D$, and $k_1$ and $b$ are two parameters
(typically set to 1.2 and 0.75, respectively \cite{Robertson09}).
Leaving aside the mathematical details for the moment being, the
important thing to note is that the right-hand side of both equations
consists of the product of two factors, i.e., a
\emph{document-dependent} factor (the leftmost one), which depends on
the term's frequency in the document \cite{luhn1957statistical}, and a
\emph{collection-dependent} factor (the rightmost one), which depends
on the rarity / specificity of the term in the collection
\cite{Sparck-Jones:1972fk}.  Accordingly, many term weighting
functions $\tw$ share the common structure
\begin{equation}
  \tw(t,d,D,C)=\dd(t,d)\cdot \cd(t,D,C)
  \label{eq:tfidflike}
\end{equation}
\noindent where $\dd(t,d)$ and $\cd(t,D,C)$ are two ``abstract'',
generic functions representing the document-dependent and the
collection-dependent factors, respectively\footnote{The $C$ parameter
in $\tw(t,d,D,C)$ and $\cd(t,D,C)$ represents the set of classes, and
is thus relevant only in text classification (or related) contexts.
While functions used in \textit{ad hoc} retrieval, such as those of
Equations \ref{eq:logtfidf} and \ref{eq:bm25}, do not depend on it, we
include this parameter because it will prove useful when discussing
instantiations of Equation \ref{eq:tfidflike} in text classification
contexts.}.
% which clearly stands there is a directly proportional
% % trade-off
% dependency between the frequency of a term in a document and its
% rarity in the collection.  This is computed through the
% $\mathit{tf}$ function, that locally computes a score based on the
% term frequency in the document, and the $\mathit{idf}$ function,
% that globally computes a score based on the term frequency in the
% collection.
Aside from these two factors, in many term weighting functions there
is a third component in the weighting process, i.e., the
\emph{normalisation component} \cite{Na:2015ve,Singhal96,Zobel98},
whose aim is to factor out document length. In variants of Equation
\ref{eq:logtfidf} it often takes the form of a denominator that
normalises the entire function, e.g., implementing cosine
normalisation; in Equation \ref{eq:bm25}, instead, a normalisation
component is already present in the denominator of the first factor.
% \fabscomment{Qui ce la caviamo troppo comodamente, sarebbe bello
% avere un approccio dove tutte e tre le componenti (tf, idf,
% normalizzazione) possono essere apprese ``jointly''.}

The weighting functions of Equations \ref{eq:logtfidf} and
\ref{eq:bm25} are \emph{unsupervised}, i.e., they do not leverage the
class labels of the training data when these latter are available.
However, in tasks (such as text classification) characterised by the
presence of training data it seems logical to think that one could
take advantage of the information contained in these labels; this is
the principle at the basis of \emph{Supervised Term Weighting} (STW)
\cite{SAC03b}. The main intuition that underlies STW is that, in the
presence of training data, the $\cd(t,D,C)$ factor in Equation
\ref{eq:tfidflike}, which measures the discrimination power of term
$t$ independently of any particular document $d$, can be instantiated
by means of a function that measures such discrimination power in a
manner more specific to the task at hand (e.g., classification). In
other words, while a traditional unsupervised $\cd(t,D,C)$ function
reflects the distribution of $t$ in the collection (and does not
depend on the class labels $C$), STW replaces it with a function that
reflects this distribution \emph{conditioned on} $C$; in this way, a
higher weight is thus given to those terms whose distribution in the
training data is better correlated with the distribution of the
labels.
% Supervised Term Weighting (STW) \cite{debole2004supervised} consists
% therefore in calculating, for each term in a document, a score value
% reflecting the relative importance of the term to the document, by
% taking into consideration the term frequency and the document
% classes in the training set.

Several metrics could be directly borrowed from the literature as
candidates for this supervised $\cd(t,D,C)$ factor.  Of particular
interest are the feature-scoring functions (hereafter: ``FS
functions'') used in ``filter-style'' feature selection
\cite{John94,Yang97}.  Previous work in STW
\cite{Batal:2009uq,SAC03b,Soucy:2005fk,Lan:2009kx,quan2011term} has
focused on identifying a FS function that performs well when used to
instantiate the $\cd(t,D,C)$ factor in Equation \ref{eq:tfidflike};
the most widely used such functions include \emph{chi-square} (denoted
$\chi^2$), which evaluates the mutual statistical dependence of two
random variables, and \emph{information gain} (a.k.a.\ \emph{mutual
information}, see Equation \ref{eq:chisquare}), which measures the
reduction in the entropy of one random variable (here: the class
label) caused by the observation of another variable (here: the
presence/absence of the term).

In STW, the $\cd(t,D,C)$ component of Equation \ref{eq:tfidflike} is
instantiated by one of these FS functions (e.g., information gain);
unlike the second factors of Equations \ref{eq:logtfidf} and
\ref{eq:bm25}, these feature-scoring functions do depend on $C$, i.e.,
on the set of classes of interest. In this article we will restrict
our attention to the binary classification case, where
$C=\{c,\overline{c}\}$ \bblue{\label{ledueclassi}(with $\overline{c}$
indicating the complement of class $c$)}; we leave the extension to
the multiclass case (i.e., $|C|>2$) to future work.
% ; we thus represent $C$ as a binary $|D|$-dimensional vector where 1
% and 0 indicate that the document is a positive or a negative example
% of $c$, respectively.  \fabscomment{Questa frase forse qui non è il
% posto giusto dove metterla; come si chiama questo vettore nel resto
% del papero?}

Despite the fact that all STW methods (a) do have an intuitive basis,
and (b) have shown some success in empirical evaluations, there are no
clear indications that one of them is consistently superior to the
others. This means that a system designer has to resort to a
trial-and-error method on a validation set in order to select the
best-performing weighting criterion for the application of interest.
% , which sometimes turns out to be an unsupervised term weighting
% approach after all.
This absence of a clear winner in the supervised term weighting camp
may indicate that STW is yet to be fully understood; we attempt to
provide some deeper insight on the nature of STW in Section
\ref{sec:analysis}.
% and motivate our method on what we think to be more reliable
% assumptions.

% What seems to be rather clear though, is that the key for success in
% term weighting rests on the design of the $\cd(t,D,C)$ factor (while
% the influence of the term frequency is less open to
% interpretations).

% , while the $\dd(t,d)$ factor has remained unaltered.}
In this article we propose a STW framework for text classification
that learns the $\cd(t,D,C)$ function from the training data; we call
this framework \emph{Learning to Weight} (LTW).
% LTW thus \fabscomment{(unclear advantage)} avoids the costly
% exploration of candidate term weighting functions on held-out data.
The rationale behind LTW is that past work on term weighting (either
unsupervised or supervised) suggests that whether a specific
$\cd(t,D,C)$ function is optimal or not is data-dependent, which means
that it may be better to directly learn the optimal function from the
data.
% \fabscomment{La giustificazione è un po' debole: un algorithmo può
% funzionare meglio su un dataset e un altro algoritmo può funzionare
% meglio su un altro, questo vale per ogni algoritmo in ogni
% settore. In particolare, nel Machine Learning c'è il famoso ``No
% Free Lunch Theorem'', che più o meno dice che non esiste un
% algoritmo che sia il migliore in ogni applicazione e su ogni
% dataset.}

We propose various instantiations of this framework, each of them
relying on neural network models
% \fabscomment{``neural network models'' o ``neural networks''?}
that learn this function via optimization.  It is important to remark
that
% that neural networks will not be used here in the ``conventional''
% manner. That is,
our goal here is \emph{not} producing word embeddings or dense
representations of documents. Instead, we resort to neural networks as
a means of learning the optimal STW function that is to be applied to
each (nonzero) element of a sparse vector of term frequencies.  In
order to gain generality, the optimization process we propose is
independent of the learning algorithm used for training the classifier
(and of the loss minimized by this algorithm), and is instead based on
solving the simpler auxiliary problem of improving the linear
separation between the positive and negative examples.  The
experiments we have conducted show that our method outperforms
previous term weighting approaches for text classification.
Furthermore, our exploration of the learned function brings about some
interesting insights on the geometrical shape of the ``ideal''
$\cd(t,D,C)$ function.

The remainder of the article is structured as follows.  Section
\ref{sec:relatedwork} reviews previous work on supervised term
weighting, while Section \ref{sec:analysis} analyses the problems
inherent in term weighting for text classification.  Section
\ref{sec:method} presents our LTW approach, followed by Section
\ref{sec:experiments} in which we discuss the experimental evaluation
we have carried out.  \bblue{In Section \ref{sub:learntf} we discuss
whether it might make sense to also learn, aside from the $\cd(t,D,C)$
factor, also the $\dd(t,d)$ factor.} Section \ref{sec:conclusions}
concludes and outlines possible avenues for future
research. \fabscomment{Checked, Aug 14.}

% ---------------------------------------------

\section{Related Work}
\label{sec:relatedwork}

\noindent \bblue{The history of term weighting goes back to the
earliest vector-based models and probabilistic models of IR, which
were developed in the '60s (see e.g., \cite[\S
3]{Baeza-Yates:2011xd}). Two contributions that have withstood the
test of time, and that form the basis of nowadays' term weighting
functions, are the two notions that we have already discussed in
Section \ref{sec:introduction}, i.e., \emph{term frequency} (whose
origins can be traced back to the development of the SMART system
\cite{Salton:1971ye}) and \emph{inverse document frequency} (which was
first formalized in \cite{Sparck-Jones:1972fk}). Many variants of
Equations \ref{eq:logtfidf} and \ref{eq:bm25}, which both combine the
two notions in one single formula, have been developed over the years
and tested against each other (for two large-scale comparisons see
\cite{Salton88,Zobel98}). While these two notions were originally
devised for text search, over the years they have been adopted in an
essentially unchanged form for other text mining tasks, such as text
clustering (see e.g., \cite{Whissell:2011hw}) and text classification
(see e.g., \cite[\S 5.1]{ACMCS02}).}

\textbf{Supervised term weighting}.  \bblue{While this ``acritical''
adoption seems justified for clustering, which is unsupervised in
nature, it seems not for classification, where additional information
useful for weighting purposes can be extracted from training
examples. This idea is at the heart of supervised term weighting.} STW
goes back to 2003, when Debole and Sebastiani \cite{SAC03b} proposed
to reuse the scores obtained from FS functions during (supervised)
feature selection, in order to instantiate the $\cd(t,D,C)$ factor of
Equation \ref{eq:tfidflike}, i.e., as a substitute of the more
conventional $\log\frac{|D|}{n}$ and
$\log{\left(\frac{|D|-n+0.5}{n+0.5}\right)}$ factors of Equations
\ref{eq:logtfidf} and \ref{eq:bm25}.  The three variants investigated
in that work used chi-square (Equation \ref{eq:chisquare}),
information gain (Equation \ref{eq:ig}), and gain ratio (Equation
\ref{eq:gr}) as instances of $\cd(t,D,C)$, i.e.,
%%
% \begin{equation}
%  \chi^2(t,D,C)=\frac{|D|(\Pr(t,c)\cdot \Pr(\overline{t},\overline{c}) - \Pr(\overline{t},c)\cdot \Pr(t,\overline{c}))^2}{\Pr(t) \cdot \Pr(\overline{t}) \cdot \Pr(c) \cdot \Pr(\overline{c})}
%  \label{eq:chisquare}
%\end{equation}
%%
%\begin{equation}
%  \ig(t,D,C)= 
%  \sum_{t'\in\{t, \overline{t}\}}\sum_{c'\in\{c, \overline{c}\}}\Pr(t',c')\log{\frac{\Pr(t',c')}{\Pr(t')\Pr(c')}}
%  \label{eq:ig}
%\end{equation}
%%
%\begin{equation}
%  \gr(t,D,C)= 
%  \frac{\ig(t,D,C)}{-\sum_{c'\in\{c, \overline{c}\}} \Pr(c')\log_2{\Pr(c')}}
%  \label{eq:gr}
%\end{equation}
%%
%
\begin{eqnarray}
  \hspace{-5ex}\chi^2(t,D,C) \hspace{-1ex} & = & \hspace{-1ex} \frac{|D|(\Pr(t,c)\cdot \Pr(\overline{t},\overline{c}) - \Pr(\overline{t},c)\cdot \Pr(t,\overline{c}))^2}{\Pr(t) \cdot \Pr(\overline{t}) \cdot \Pr(c) \cdot \Pr(\overline{c})}
                                                 \label{eq:chisquare} \\
  \hspace{-5ex}\ig(t,D,C) \hspace{-1ex} & = & \hspace{-1ex} 
                                              \sum_{t'\in\{t, \overline{t}\}}\sum_{c'\in\{c, \overline{c}\}}\Pr(t',c')\log{\frac{\Pr(t',c')}{\Pr(t')\Pr(c')}}
                                              \label{eq:ig} \\
  \hspace{-5ex}\gr(t,D,C) \hspace{-1ex} & = &  \hspace{-1ex} 
                                              \frac{\ig(t,D,C)}{-\sum_{c'\in\{c, \overline{c}\}} \Pr(c')\log_2{\Pr(c')}}
                                              \label{eq:gr}
\end{eqnarray}
\noindent where $C=\{c,\overline{c}\}$ and $\Pr$ denotes a probability
on the event space of documents (e.g., $\Pr(\overline{t},c)$ is the
probability that a random training document does not contain $t$ and
belongs to class $c$).  However, their results (using Support Vector
Machines as the learning algorithm) did not show STW to bring about
consistent improvements over TFIDF.

A subsequently proposed STW approach is \emph{ConfWeight}
\cite{Soucy:2005fk}, which adopts a relevance criterion based on
statistical confidence intervals; the criterion can simultaneously be
used for performing feature selection \emph{and} term weighting. At
the same time, \cite{Lan:2005uq} proposed \emph{relevance frequency},
refined in \cite{Lan:2009kx} as
% ; and then refined in \cite{Lan:2009kx},} as
%
\begin{equation}
  \rf(t,D,C)=\log\left(2+\frac{\Pr(t,c)}{\max\{\frac{1}{|D|},\Pr(t,\overline{c})\}}\right)
  \label{eq:rf}
\end{equation}
%
% \noindent where $TP$ and $FP$ stand for the number of true positives
% and false positives, respectively, in the binary contingency table
% (e.g., $FP$ is the number of training documents in which $t$ occurs
% and which do not belong to $c$).

% Those counters, along with the $FN$ and $TN$ (the \emph{false
% negatives} and \emph{true negatives}, respectively) come from the
% 4-cell contingency table (Table \ref{eq:4cell}) that accounts for
% the frequency distribution of the discrete variables $t$ and $c$ in
% terms of presence/absence.
% %
% \begin{table}[t!]
%   \centering
%   \begin{tabular}{c|cc}
%     & $c$ & $\overline{c}$ \\ \hline
%     $t$            & TP  & FP             \\
%     $\overline{t}$ & FN  & TN           
%   \end{tabular}
%   \caption{4-cell contingency table for a \emph{decision stump} term $t$ with respect to the true outcomes $c$.}
%   \label{eq:4cell}
% \end{table}
% Note that all equations above discussed are liable to be expressed
% as functions of the contingency table, since all probabilities from
% Equations \ref{eq:chisquare} to \ref{eq:gr} are directly obtainable
% from this table.

\noindent as the instance of the $\cd(t,D,C)$ factor. STW based on
relevance frequency (which we use as a baseline in the experiments of
Section \ref{sec:experiments}) was found by \cite{Lan:2005uq} to be
superior to STW using $\chi^{2}$ or $\ig$, but comparable to standard
TFIDF under certain circumstances.

Experiments similar to the ones of Debole and Sebastiani \cite{SAC03b}
were carried out in \cite{Batal:2009uq}, where instead STW proved
useful for improving the classification accuracy of $k$-Nearest
Neighbours (KNN) classifiers.

Other variations on the same theme are to be found in
\cite{Domeniconi2015,Haddoud2016,Shanavas2016,Wang2013,Wu2017}, while
applications of STW to specific instantiations of text classification,
such as question classification \cite{quan2011term} or sentiment
classification \cite{Kim2014, Nguyen2011}, have also been reported.

Overall, the results reported in the literature show, as hinted above,
that there is no clear consensus as which variant of STW is the best,
and as to whether STW is indeed superior or not to standard
unsupervised weighting.

% Separability index (notes) \cite{mthembu2008note}

% First notion of tf-like \cite{luhn1957statistical}

% First notion of idf-like \cite{Sparck-Jones:1972fk}

% Binary:
% \begin{equation}
%binary(t,d)=\left\{\begin{matrix}
%1 :  t \in d \\ 
%0 : t \notin d
% \end{matrix}\right.
%\label{eq:binary}
%\end{equation}

%http://kak.tx0.org/IR/TFxIDF
%\begin{equation}
%BM25(t,d,D) = \frac{(k_1+1)\cdot tf}{k_1\left((1-b)+b\cdot\frac{dl}{avdl}\right)+tf}\cdot\log{\left(\frac{N-n+0.5}{n+0.5}\right)}
%\label{eq:bm25}
%\end{equation}
%Where $k_1=1.2$, and $b=0.75$

%\begin{equation}
%tfidf(t,d,D) = \left(1+\log(f_{dt})\right)\cdot\log\frac{|D|}{|\{d \in D : t \in d\}|}
%\label{eq:logtfidf}
%\end{equation}

\textbf{Representation learning}. Term weighting document vectors is
inherently related to \emph{representation learning}, an area of
research where neural networks, and in particular deep learning
architectures \cite{lecun2015deep}, tend to outperform the
competition.  Deep learning allows computational models composed of
non-linear projections to learn effective representations of the
inputs by backpropagating the errors with respect to the model
parameters.  In recent work \cite{Mikolov:2013jk} deep learning models
have been used to obtain continuous, dense representations for words
and document vectors.  Such methods have proved effective at modelling
word semantics, but require the processing of very large external text
collections in order to succeed.  Bag-of-words approaches to text
classification, which are the context of our LTW work, do not require
any external text collection.  In the experiments we include a
comparison with \textsc{FastText}
\cite{Bojanowski:2017fe,Grave:2017wj}, a top-performing method based
on distributional semantics that uses dense representations.
% We are instead interested in preserving data sparsity, because this
% allows much higher levels of efficiency at both training and
% classification time.

In the present work neural networks are not used for training a text
classifier (for this, a neural network or any other learning algorithm
could be used); instead, the rationale of using neural networks is (a)
to exploit their modelling flexibility in order to improve the
weighting criterion, %of document vectors,
and (b) to permit the inspection of the learned weighting function, in
order to allow the experimenter to gain intuitions on how an ideal
such function looks like.

\textbf{Learning term weights.}. To the best of our knowledge, the
\bblue{\emph{Combined Component Approach} (CCA)} is the only previous
approach resembling the idea of learning term weighting functions. CCA
was presented in \cite{de2007combined}, in the context of learning to
rank.
% \fabscomment{C'erano anche i lavori di Cummins e O'Riordan, e.g.,
% \cite{Cummins2006, Cummins2006a}; se anche loro fanno la stessa
% cosa, tanto vale citare anche loro.}  \fabscomment{Mi sa che Cummins
% e O'Riordan, dato che lavorano in ambito ad hoc search, le label
% proprio non le hanno.}  The latter paper
CCA follows a radically different approach, though, based on
composing, via genetic programming optimization, complex ranking
functions from 20 weighting factors well-known from past literature
(e.g., \emph{tf} and \emph{idf} variants, among others) used as the
atomic components. Differently from \cite{de2007combined}, instead of
exploring the (huge -- see \cite{Zobel98}) space of combinations of
previously proposed weighting factors, we take the simpler \emph{tpr}
and \emph{fpr} statistics (see Section \ref{sec:method}) as the atomic
components, and allow the optimization procedure full freedom. The
main difference between CCA
% the method of \cite{de2007combined}
and our method is thus the fact that we use the supervision directly
as input to the weighting function learning method (in order to
compute the \emph{tpr} and \emph{fpr} statistics), while
CCA %\cite{de2007combined}
only uses the supervision in the evaluation phase, i.e., their
optimization procedure works primarily by composing unsupervised
weighting factors. \bblue{We include CCA as one of our baselines in
the experiments of Section \ref{sec:experiments}.}
% Thanks to the current capabilities of modern GPUs, we found our
% algorithm converges in less than a minute in average. }\alex{Loro
% non specificano niente sul tempo di convergenza, ma essendo un GP
% direi che e' piu' tosto lento.}

% ---------------------------------------------

\section{Pitfalls in Supervised Term Weighting Functions}
\label{sec:analysis}

\noindent
% \fabscomment{Forse questa sezione andava chiamata ``Open issues in
% Supervised Term Weighting functions''? Il titolo attuale mi sembra
% ... un po' ambizioso.}
In this section we analyse some of the typical instantiations of the
components of supervised term weighting functions, trying to shed some
light on the possible reasons why there is no consensus yet on which
among the many available variants is preferable.

% ---------------------------------------------

\subsection{The $\dd(t,d)$ Factor}
\label{sec:analysis-tf}

\noindent The most frequently used unsupervised variants of the
$\dd(t,d)$ factor of Equation \ref{eq:tfidflike} include the raw
frequency of $t$ in $d$ (Equation \ref{eq:rawfreq}) and log-scaled
versions of it (Equations \ref{eq:lognorm} and \ref{eq:lognorm2}):
% \fabscomment{Di queste due forse varrebbe la pena lasciare solo la
% seconda; la prima davvero non ha senso!!!}: (Although other variants
% of this factor exist in the literature we will not consider them in
% this discussion as they are rarely used in practice.)
% \andscomment{L'ho commentato, al massimo lo metterei come footnote.}
%
\begin{eqnarray}
  \label{eq:rawfreq}
  \dd(t,d) & = & f_{td} \\ 
  \label{eq:lognorm}
  \dd(t,d) & = & \left\{ \begin{array}{ll} 
                           1+\log
                           f_{td} & \mbox{if } f_{td}>0 \\ 0 & \mbox{otherwise}
                         \end{array} \right . \\
  \label{eq:lognorm2}
  \dd(t,d) & = & \log \left(1+f_{td} \right)
\end{eqnarray}
\noindent In this paper we focus on the $\cd(t,D,C)$ function; this
allows us to draw a fair comparison with existing STW methods, which
do not address the $\dd(t,d)$ component of the weighting function.
However, in this section we would like to discuss a few minor issues
that affect traditional implementations of the $\dd(t,d)$ function.

The linear version of Equation \ref{eq:rawfreq} does not take into
account the fact that a term frequency variation at low frequency
values, e.g., from 1 to 3, should be considered much more relevant
than a term frequency variation at high frequency values, e.g, from 11
to 13.  For this reason log-scaled versions such as those of Equations
\ref{eq:lognorm} and \ref{eq:lognorm2} are usually preferred, since
they better capture frequency variations observed at low frequency
values.

However, the non-linearity of log-scaled versions causes issues with
document length.  To illustrate this, let us consider a document
$d\in D$ and a document $d'\in D$ which consists of two juxtaposed
copies of $d$.  The value of $f_{td'}$ is double the value of
$f_{td}$; since the log transformation is non-linear, if we use
Equation \ref{eq:lognorm} or Equation \ref{eq:lognorm2} the
cosine-normalised vectors of $d$ and $d'$ will not be the same,
potentially clashing with the so-called ``normalisation assumption'',
according to which ``for the same quantity of term matching, long
documents are no more important than short documents'' \cite{Zobel98}.

\subsection{The $\cd(t,D,C)$ Factor}
\label{sec:idflikefactor}

\noindent FS functions have proven useful in filter-style approaches
to feature selection \cite{Yang97}.  Feature selection and term
weighting are inherently related, as both tasks build upon a model of
feature importance, which is what FS functions aim to measure.  One
might thus expect FS functions to fit the purposes of term weighting
(an intuition that has driven a lot of previous work in STW -- see
Section \ref{sec:relatedwork}). However, there are reasons to believe
that a good FS function is not necessarily a good $\cd(t,D,C)$
component of a term weighting function. The reason is that our notion
of the quality of a FS function is typically based on its performance
as a ranking function, since FS functions are customarily used in
filter-style feature selection in order to rank features.  That is, if
$fs$ is a FS function measuring the degree of importance of a feature
(whatever this might mean), filter-style feature selection consists of
taking the top $n$ ranked features according to their $fs$ score.

\begin{figure}[h!]
  % \centering
  \includegraphics[height=20.8cm]{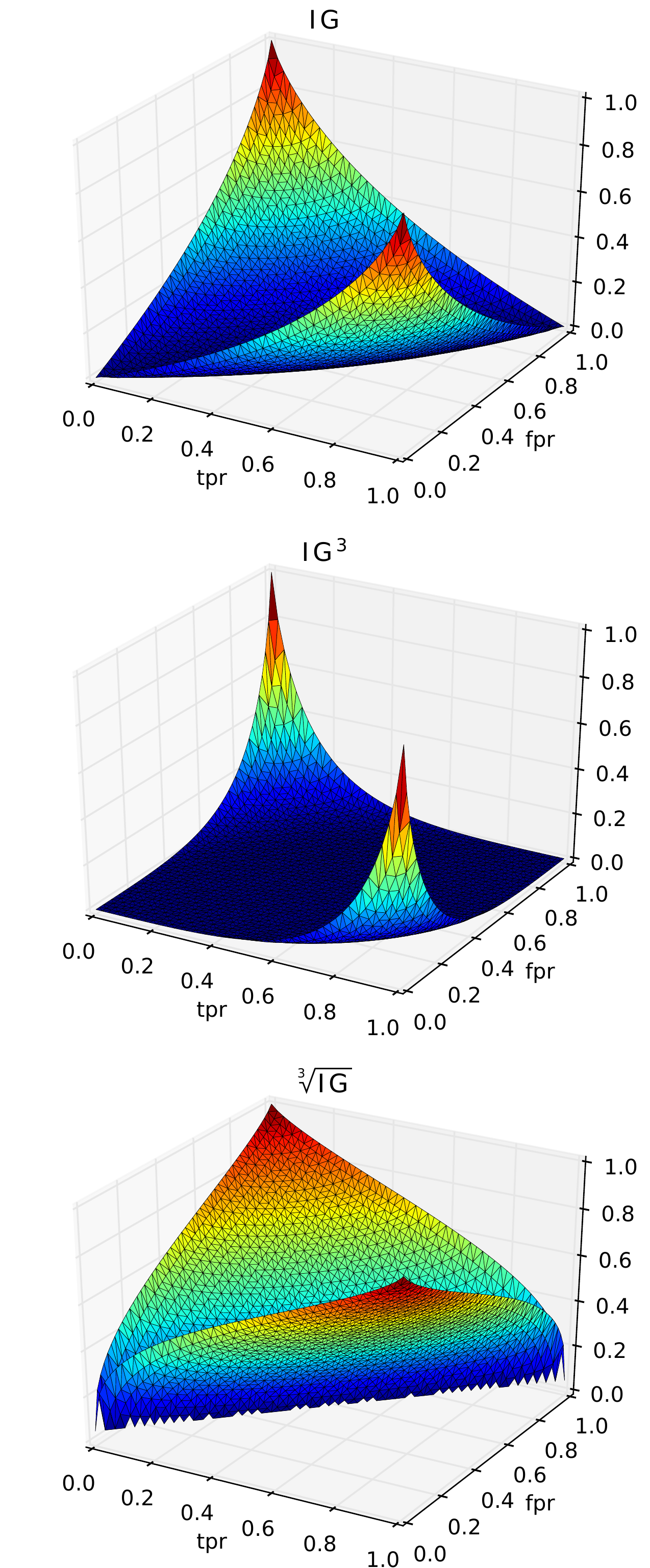}
  \caption{Plots, from top to bottom, include information gain ($\ig$)
  unaltered, information gain cubed, and the cubed root of information
  gain as a function of the true positive ratio $tpr$ and the false
  positive ratio $fpr$ (explained in detail below). Despite being very
  dissimilar when used as weighting criteria, the three versions of
  $IG$ produce identical feature rankings when used as FS functions.}
  \label{fig:fs_as_stw}
\end{figure}

Note that the numeric values produced by a FS function can be
substantially modified by applying to them any monotonic
non-decreasing function, without affecting the resulting rankings.
For example, the three variants of the $\ig$ function in Figure
\ref{fig:fs_as_stw} produce exactly the same results when used to rank
features, but can result in very different outcomes when used to
instantiate the $\cd(t,D,C)$ function.  This indicates that, when used
in a supervised term weighting formula, for any FS function there is
an additional dimension to be explored for optimization, i.e.,
monotonic non-decreasing transformations.  Instead of systematically
exploring the space of possible such transformations we propose to
learn the weighting function from scratch on the training set, without
relying on predefined term relevance functions. This proposal will be
detailed in the next section.

% Inspecting the variants (logs, logs+1, normalisations factors such
% as gain ratio, parameters of BM25, etc.). One can not pretend to
% know which is preferable in advance. Move to a learnt function,
% adjusted for each dataset, and aware of normalisation.
% ---------------------------------------------

\section{Learning to Weight}
\label{sec:method}

\noindent We propose a novel supervised weighting approach that,
instead of relying on any predefined formula, learns a term weighting
function optimised on the available training set; we call this
approach \emph{Learning to Weight} (LTW)\footnote{We should remark
that we do \emph{not} optimise the weighting function for a specific
\emph{test} set. The fact that we do not use the test set in any way
while learning the weighting function sets our approach apart from the
realm of \emph{transductive} learning, and squarely places it within
the domain of \emph{inductive} learning.}. Essentially, during
training, different term weighting functions (leading to different
term weights for the same term-document pair) are tested, and the
function that minimizes the linear separation between the positive and
negative examples of the training set is chosen. The different term
weighting functions that are tested are generated by neural
architectures described in Section \ref{sec:learningtheidflike}.

\begin{figure}[h!]
  % \centering
  \includegraphics[height=22.0cm]{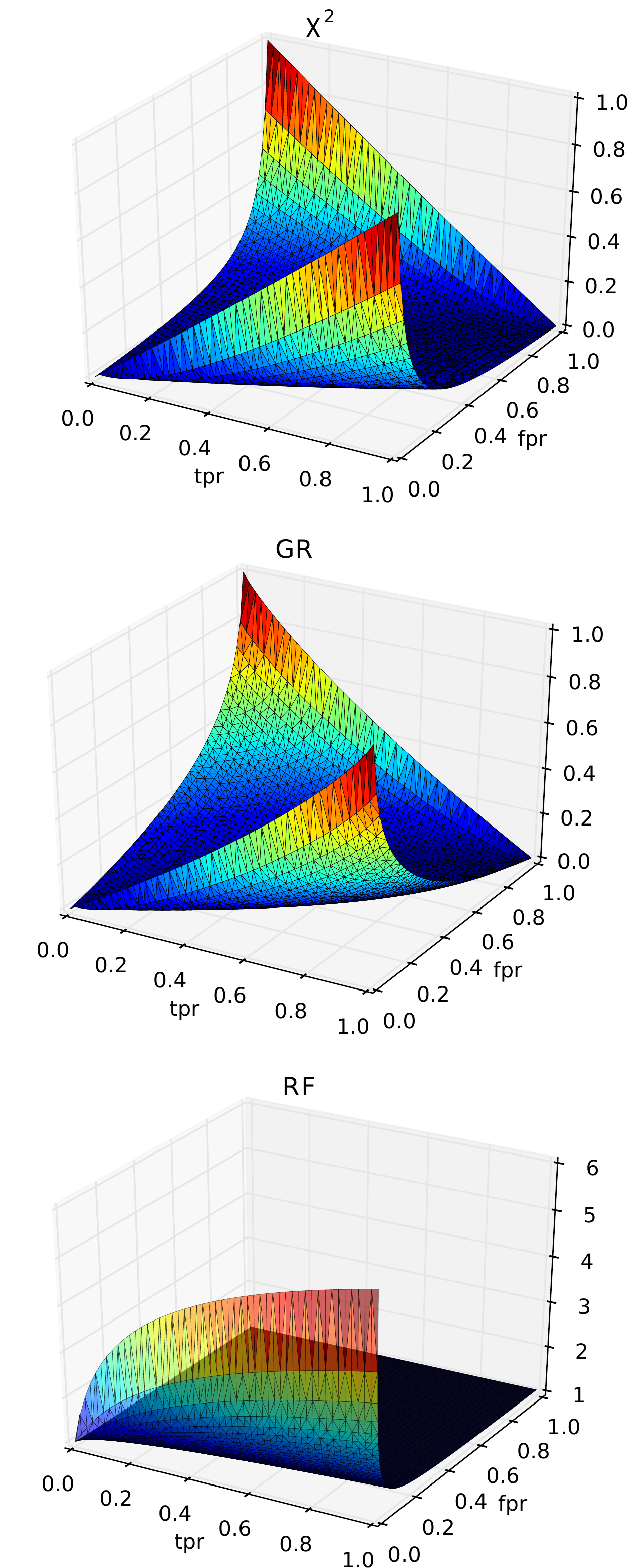}
  \caption{Plots, from top to bottom, of chi-square ($\chi^2$), gain
  ratio (GR), and relevance frequency (RF), as a function of the true
  positive ratio $tpr$ and the false positive ratio $fpr$.  All plots
  were generated assuming $\Pr(c)=0.05$.}
  \label{fig:chi_gr_rf}
\end{figure}

In the rest of the paper we will use slightly different mathematical
conventions from the ones used in Section \ref{sec:relatedwork}. In
fact, while the four functions discussed in that section are all
formulated in terms of the four variables
$\{\Pr(t,c), \Pr(\overline{t},c), \Pr(t,\overline{c}),
\Pr(\overline{t},\overline{c})\}$, in reality they are functions of
just \emph{two} free variables, since the frequency of the class
$\Pr(c)\equiv \Pr(t,c)+\Pr(\overline{t},c)$ in the training set is
fixed, and since $\Pr(\overline{t},c) = \Pr(c) - \Pr(t,c)$ and
$\Pr(t,\overline{c}) = (1-\Pr(c)) -
\Pr(\overline{t},\overline{c})$. In the rest of the paper we will thus
express our functions in terms of just two free variables; however,
consistently with the tradition of ROC analysis \cite{Fawcett:2006fk},
as our two free variables we will equivalently choose the \emph{true
positive rate}
\begin{equation}
  \tpr(t,D,C) = \frac{\Pr(t,c)}{\Pr(c)}
  \label{eq:tpr}
\end{equation}
\noindent and the \emph{false positive rate}
\begin{equation}
  \fpr(t,D,C) = \frac{\Pr(t,\overline{c})}{\Pr(\overline{c})}
  \label{eq:fpr}
\end{equation}
\noindent As a side effect, this change of variables makes it possible
to plot any $\cd(t,D,C)$ function in 3D space, which allows the
experimenter to quickly gain an understanding on how the weights are
assigned (something that will prove useful in our analysis of the
results in Section \ref{sec:experiments}). Figure \ref{fig:chi_gr_rf}
plots the FS functions discussed in Section \ref{sec:relatedwork}
% as instances of the $\cd(t,D,C)$ factor.
as a function of the two variables $\tpr$ and $\fpr$ of Equations
\ref{eq:tpr} and \ref{eq:fpr}; note that we have not plotted
$\ig(t,D,C)$ since it differs from $\gr(t,D,C)$ by a fixed
multiplicative factor only.  As expected, the plots reveal a consensus
on the higher importance of positively correlated terms (those
characterized by high $\tpr$ and low $\fpr$). Interestingly enough,
there are instead discrepancies among these functions on (i) how to
assess the importance of negatively correlated terms (those with low
$\tpr$ and high $\fpr$), which are considered as important as the
positively correlated ones by functions $\chi^2$ and $\gr$ but ignored
by $\rf$; or on (ii) how much the importance of the term changes due
to variations in $\tpr$ and $\fpr$. \fabscomment{Checked, Aug 14.}

\subsection{Learning the $\cd(t,D,C)$ Factor}
\label{sec:learningtheidflike}

% FS functions take the 4 values of the contingency table as input to
% compute the relevancy score.
\noindent In this work we will learn a weighting function individually
for each binary classification problem.
% (and leave the case in which the whole multi-label collection is
% weighted for future research).
As discussed in Section \ref{sec:relatedwork}, in the binary case any
function of a contingency table can be expressed as a function of the
two values $\tpr(t,D,C)$ and $\fpr(t,D,C)$.  Our working hypothesis is
that the optimal such function is data-dependent. We thus propose to
use a neural network to learn this function from data, since neural
networks are universal approximators \cite{Hornik1991}.

Concretely, we adopt a multi-layer feedforward network with one hidden
layer, since this architecture, despite being simple\footnote{As part
of this framework, more complex architectures could be adopted as
well; however, this goes beyond the scope of this work (for which the
expressive power of the adopted network is sufficient) and is
something we plan to investigate in the future.}, is known to be able
to approximate any continuous function if enough hidden nodes are
considered \cite{Hornik1991}.

The $\cd(t,D,C)$ computation we propose\footnote{Note the difference
in notation between the $\cd(t,D,C)$ function of Equation
\ref{eq:tfidflike} and the $cd(t,D,C)$ functions of Equations
\ref{eq:local} and \ref{eq:global}; the former is an ``abstract''
function that measures the importance of term $t$ for dataset $D$ and
set of classes $C$, while the latter are concrete instantiations of it
that we use in this paper.  Similarly for the $\dd(t,d)$ function of
Equation \ref{eq:tfidflike} and the $dd(t,d)$ function of Equation
\ref{eq:mod-tf-like}.} is described by the following equations:
\begin{equation}
  \begin{aligned}
    & h_t  = \ReLU\left([tpr_t,fpr_t]\times W_1+b_1\right) \\
    & cd_L(t,D,C) = \sigma \left(h_t \times W_2+b_2\right)
    \label{eq:local}
  \end{aligned}
\end{equation}%
\noindent where $h_t$ is the output of the hidden layer,
$\ReLU(x)=\max(x,0)$ is the Rectifier Linear Unit activation function,
$[tpr_t,fpr_t]$ is a vector of the two precomputed values
$\tpr(t,D,C)$ and $\fpr(t,D,C)$, respectively, $W_i$ and $b_i$ (for
$i\in\{1,2\}$) are the transformation matrix and the bias term, and
$\sigma(x)=1/(1+e^{-x})$ is the logistic %(sigmoidal) activation
function.  The model parameters $W_i$ and $b_i$ (for $i\in\{1,2\}$)
are shared across the model, i.e., the same set of parameters is used
in order to compute the weight of each term, in a convolution-like
manner (i.e., all $cd_L(t,D,C)$ scores are computed in parallel on
each input pair $(tpr_t,fpr_t)$ -- see the right branch of the network
in Figure \ref{fig:LTW-local}).  The subscript $L$ stands for
\emph{feature-local}, since the weight for a feature $t$ is computed
from statistics ($tpr_t$ and $fpr_t$) local to that feature.

Traditional STW approaches have only considered the local features
$tpr_t,fpr_t$ as the inputs in order to compute the $\cd(t,D,C)$
score. However, one could think of leveraging the global information
in the collection in an attempt to model the possible
inter-dependencies among features. For example, one could think of
increasing the $\cd(t,D,C)$ score for a feature which, despite not
being too strongly correlated to the class label, has the strongest
correlation among all other features. Note that such kind of
considerations remain out of reach for all feature-local methods.

We thus propose an alternative model, dubbed \emph{feature-global},
which instead uses the statistics of all features in order to compute
the $\cd(t,D,C)$ score of a given term.  This variant is described by
the set of equations
\begin{equation}
  \begin{aligned}
    \hspace{-5em}
    % & h = \ReLU([tpr_1,fpr_1,\ldots,tpr_T,fpr_T] \times \\ &
    % \hspace{7em} \times W_1+b_1) \label{eq:global} \\
    & h  = \ReLU([tpr_1,fpr_1,\ldots,tpr_T,fpr_T] \times W_1+b_1) \label{eq:global} \\
    & o  = \sigma(h\times W_2+b_2)   \\
    & cd_G(t,D,C) = o[t]
  \end{aligned}
\end{equation}
\noindent where $W_i$ and $b_i$ (for $i\in\{1,2\}$) are the
$\cd(t,D,C)$ model parameters (with the same meaning as above) and $o$
is the output vector containing all $\cd(t,D,C)$ scores; the
$\cd(t,D,C)$ for term $t$ is returned via the $[\cdot]$ operator.  In
this setup the hidden state $h$ is not the result of a separate
parallel computation for each feature, but is instead the output of a
single-step computation that takes into account all the features at
the same time (Figure \ref{fig:LTW-global}).
% Subscript $G$ stands for global, since the idf of any term is
% computed taking all statistics

\label{fourelements} Note that, as the input features for the
$\cd(t,D,C)$ function, we could opt for the four joint probabilities
$\Pr(t,c),\Pr(\overline{t},c),\Pr(t,\overline{c}),\Pr(\overline{t},\overline{c})$,
in place of the $\tpr,\fpr$ factors. However, this would unnecessarily
complicate the system, roughly doubling the number of parameters and
additionally constraining it to discover the actual degrees of freedom
in the input data. \bblue{Conversely, using both $\tpr$ and $\fpr$
(and not, say, just one of them) is essential because $\tpr$ and
$\fpr$ are altogether defined in terms of all four joint probabilities
($\tpr=\frac{\Pr(t,c)}{\Pr(t,c)+\Pr(\overline{t},c)}$ is a function of
$\Pr(t,c)$ and $\Pr(\overline{t},c)$ while
$\fpr=\frac{\Pr(t,\overline{c})}{\Pr(t,\overline{c})+\Pr(\overline{t},\overline{c})}$
is a function of $\Pr(\overline{t},\overline{c})$ and
$\Pr(t,\overline{c})$). If a method used only one of $\tpr$ and
$\fpr$, it would work with incomplete information about how $t$ are
$c$ are correlated.  }

As a final note, all $tpr_{t}$ and $fpr_{t}$ inputs for the
$\cd(t,D,C)$ function (both local and global) are calculated once, at
the beginning, and then fixed during all the optimization process.
Note thus that the only part that varies during the training process
concerns the term frequencies of each used document vector (i.e., the
input of the $\dd(t,d)$ function) which is not directly connected to
the $\cd(t,D,C)$ part. For more discussion on this see Section
\ref{sec:conclusions}.

% ---------------------------------------------

\subsection{Composing the $\tw$ score}
\label{sec:composingthetfidflike}

\noindent In this study we instantiate the $\dd(t,d)$ factor with the
function
\begin{equation}
  dd(t,d) = \log\left(1+\frac{f_{td}}{dl}\right)
  \label{eq:mod-tf-like}
\end{equation}
\noindent i.e., a log-scaling of the frequencies of terms normalized
by the document-length ($dl$). Normalizing by the document length is a
simple way to limit the variation of the input range to the neural
network, at the same time avoiding the issues related to document
length (see Section \ref{sec:analysis-tf}).

Once the $dd(t,d)$ and $cd(t,D,C)$ factors are calculated, they are
multiplied in a pointwise manner\footnote{\bblue{Note that the tensor
resulting from the $dd(t,d)$ branch has ``shape'' $[batchsize,F]$
while the tensor from $cd(t,D,C)$ has ``shape'' $[1,F]$. This mismatch
in the pointwise multiplication is resolved via ``broadcasting'', a
typical operation in any deep learning framework.}} and then
normalised via $L_2$ normalisation, i.e.,
\begin{equation}
  \mathit{lw}(t,d,D,C)=\frac{dd(t,d)\cdot cd(t,D,C)}{\sqrt{\sum_{t\in d} \left(dd(t,d)\cdot cd(t,D,C)\right)^2}}
\end{equation}
\noindent where $\mathit{lw}$ stands for \emph{learned weights} and
$cd(t,D,C)$ is either $cd_L(t,D,C)$ or $cd_G(t,D,C)$.

The $cd(t,D,C)$ model parameters are optimised to improve the linear
separability of the positive and negative examples.  For that purpose
we use a simple logistic regression model
\begin{equation}
  \hat{y}_d=\sigma(\mathit{lw}(t,d,D,C)\times W_3 + b_3)
  \label{eq:cestimate}
\end{equation}
\noindent where $\hat{y}_d$ is the model prediction.
%
% Finally, all model parameters (both the logistic regression
% parameters and the $\cd(t,D,C)$ parameters) are learned by
% backpropagating the errors.
As the loss function we use the \emph{cross-entropy}
\begin{equation}
  \begin{aligned}
    \hspace{-2ex} \mathcal{L}(y_d,\hat{y}_d) = & \ -y_d\log
    (\hat{y}_d) -(1-y_d)\log (1-\hat{y}_d)
    \label{eq:crossentropy}
  \end{aligned}
\end{equation}
\noindent between the true label $l_d\in\{c,\overline{c}\}$ mapped
into $\{0,1\}$ by means of
\begin{equation}
  y_d = 
  \left \{
    \begin{array}{ll} 
      1 & \mbox{if } l_d=c \\ 
      0 & \mbox{if } l_d=\overline{c} 
    \end{array} \right .
\end{equation}
and the prediction logits $\hat{y}_d\in(0,1)$. We use cross-entropy
since it is known to be a good differentiable model of the error for
logistic regression.

It is important to note that the logistic regression layer defined by
$W_3$ and $b_3$ has the sole purpose of propagating the constraints on
the parameters for the $\cd(t,D,C)$ model.  That is, the logistic
regressor is merely used here as an auxiliary classifier, and the real
output of the system are the parameters $W_i$ and $b_i$ (for
$i\in\{1,2\}$) of the $\cd(t,D,C)$ function; the parameters $W_3$ and
$b_3$ of the logistic regressor are set aside once the optimization
ends.  The choice of using a simple logistic regression layer is a
minimalistic one, allowing to directly correlate the values of the
$\mathit{lw}(t,d,D,C)$ function to the class labels, with minimal bias
towards the actual classifier being used. Note that resorting to a
more sophisticated classifier (e.g., a deep multi-layer feedforward
network) could cause the contribution of the supervised factor to be
diminished, as the (auxiliary) net could well end up delegating the
modelling of complex aspects of the data to the inner layers. Stacking
the simplest possible auxiliary classifier on top thus forces the
quality of the $\mathit{lw}$ layer (the actual outcome of the model)
to be maximized.
% \alex{Il $\log()$ e'sempre definito perché $\hat{c}$ (l'output della
% logistic function) non e' mai uguale a 0 ne a 1.}  The optimization
% process thus performs a gradient descent-driven search in the
% parameter space $[W_i,b_i]_{i=1,2,3}$ to minimize the averaged loss.
% can be discarded once the optimization has ended.

\begin{figure*}[tb]
  \centering
  \begin{subfigure}[b]{.9\columnwidth}
    \includegraphics[height=.8\columnwidth,left]{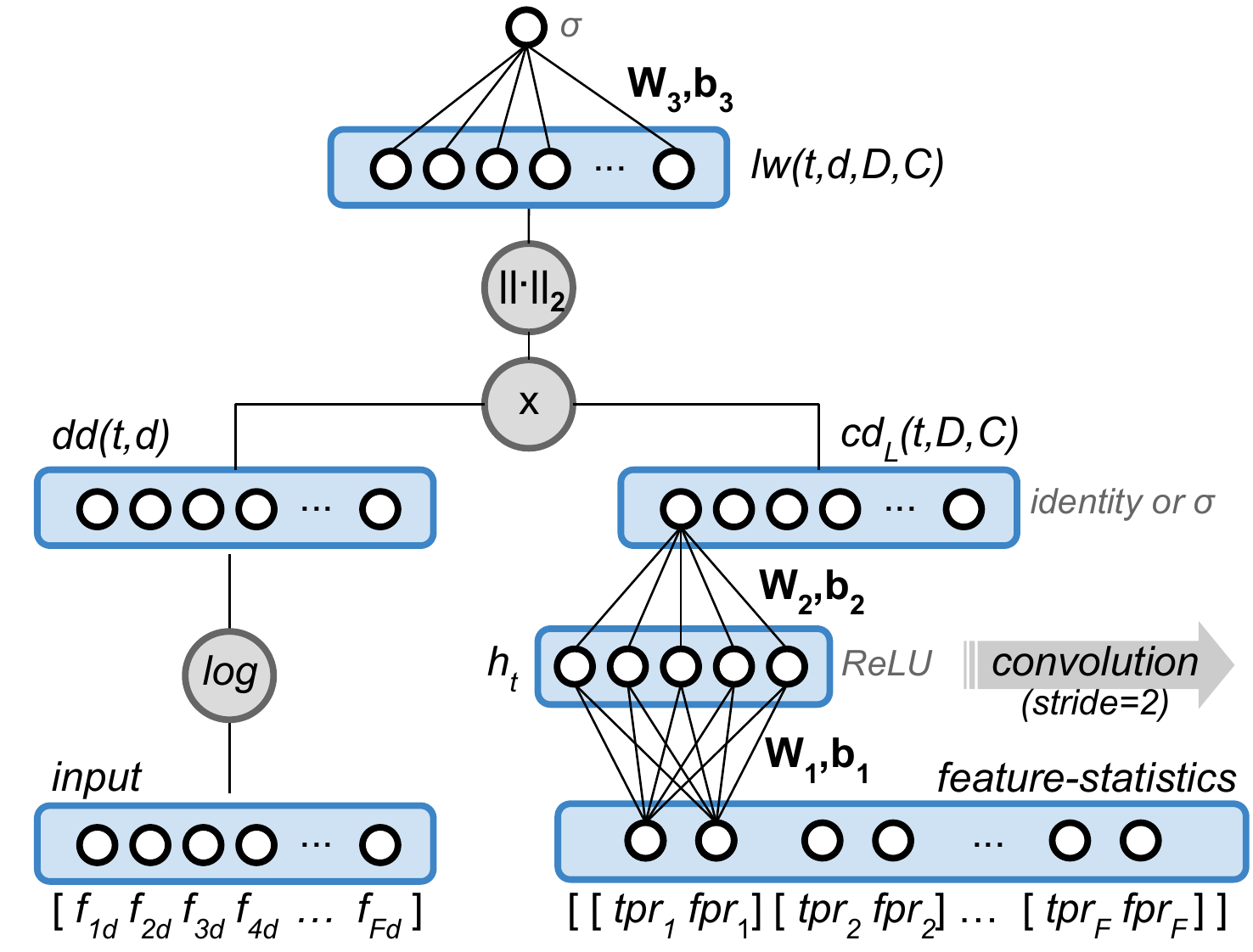}
    \caption{}\label{fig:LTW-local}
  \end{subfigure} \quad
  \begin{subfigure}[b]{\columnwidth}
    \includegraphics[height=.72\columnwidth,right]{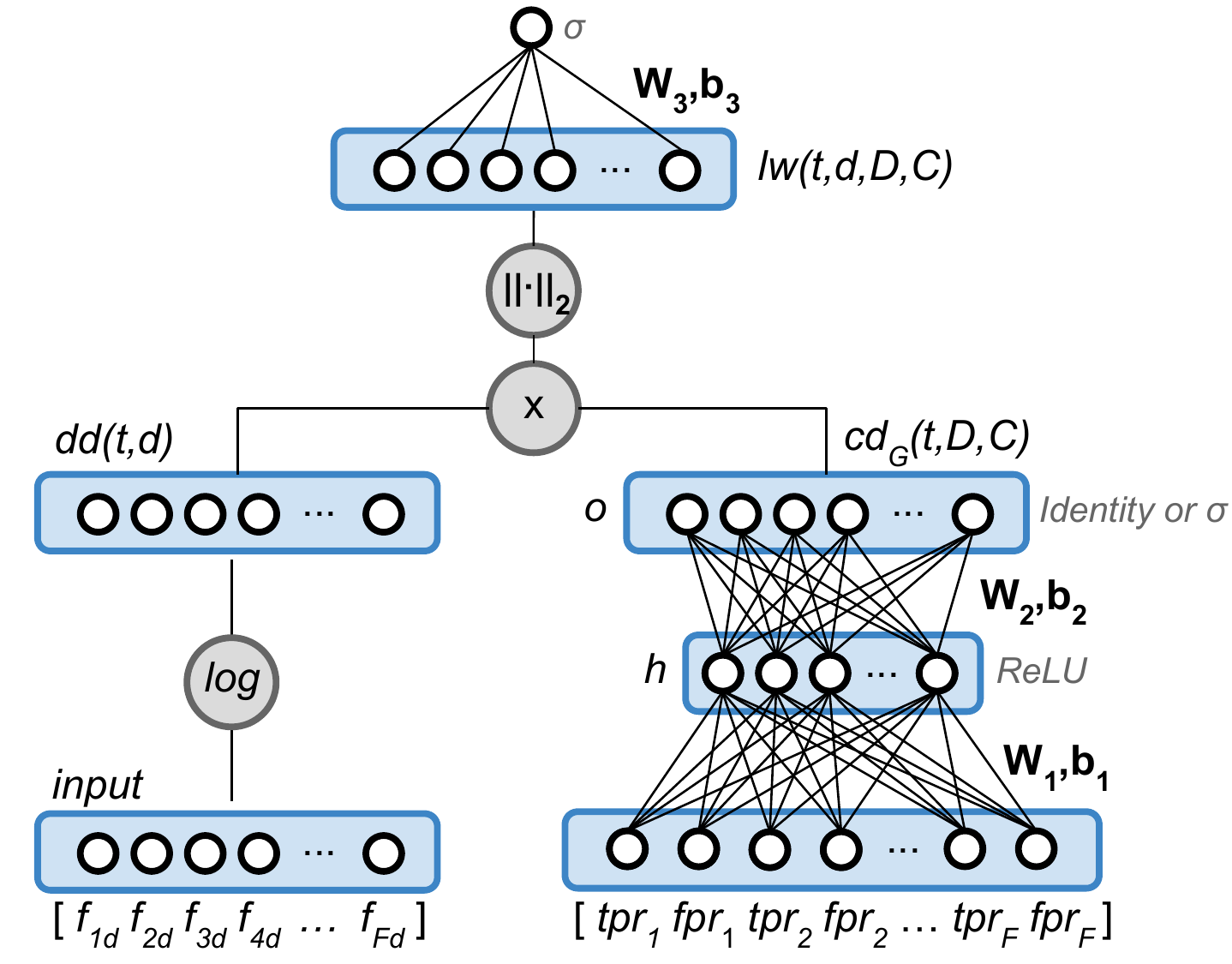}
    \caption{}\label{fig:LTW-global}
  \end{subfigure}
  \caption{Learning to Weight architectures: local variant (a), and
  global variant (b).}
  \label{fig:LTW-arch}
\end{figure*}

Figures \ref{fig:LTW-local} and \ref{fig:LTW-global} show the
computation graph in the local and global variants, respectively.

Note that the Learning to Weight framework gives full control to the
optimization process for balancing the different factors involved in
the weighting process.  For example, if a dataset can be easily
separated by exclusively looking at the frequency of its terms, then
the learning process will force the $\cd(t,D,C)$ function to mimic the
constant function $cd(t,D,C)=1$.  Otherwise, if the term frequency
adds little information, the optimization process will try to
compensate it by increasing the importance of the $\cd(t,D,C)$ factor.

% ---------------------------------

\section{Experiments}
\label{sec:experiments}

\noindent In this section we experimentally compare our Learning to
Weight framework to other supervised and unsupervised term weighting
methods proposed in the literature\footnote{\label{foot:code}The code
for reproducing all the experiments discussed in this paper is
available at \url{https://github.com/AlexMoreo/learning-to-weight}}.

% \begin{figure}[tb]
%   \centering
%   \includegraphics[height=2in]{Figures/LTW-local}
%   \caption{Learning to Weight architecture, local variant.}
%   \label{fig:LTW-local}
% \end{figure}
%%
% \begin{figure}[tb]
%   \centering
%   \includegraphics[height=2in]{Figures/LTW-global}
%   \caption{Learning to Weight architecture, global variant.}
%   \label{fig:LTW-global}
%% \end{figure}
% \begin{figure*}[tb]
%   \centering
%   \begin{minipage}[b]{.9\columnwidth}
%     \includegraphics[height=.8\columnwidth,left]{Figures/LTW-local}
%     \caption{Learning to Weight architecture, local
%     variant.}\label{fig:LTW-local}
%   \end{minipage} \quad
%   \begin{minipage}[b]{\columnwidth}
%     \includegraphics[height=.72\columnwidth,right]{Figures/LTW-global}
%     \caption{Learning to Weight architecture, global
%     variant.}\label{fig:LTW-global}
%   \end{minipage}
% \end{figure*}

% ---------------------------------------------

\subsection{Datasets}
\label{sec:datasets}

\noindent As the datasets for our experiments we use the popular
\textsc{Reuters-21578}, \textsc{20Newsgroups}, and \textsc{Ohsumed}
corpora:

\begin{itemize}

\item \textsc{Reuters-21578} is a publicly
  available\footnote{\url{http://www.daviddlewis.com/resources/testcollections/reuters21578/}}
  test collection which consists of a set of 12,902 news stories,
  partitioned (according to the ``ModApt\'e'' split we adopt) into a
  training set of 9,603 documents and a test set of 3,299
  documents. In our experiments we restrict our attention to the 115
  classes with at least one positive training example.  After removing
  stopwords, the number of distinct terms amounts to 28,828.  This
  dataset presents cases of severe imbalance, with several classes
  containing fewer than 5 positive examples.

\item \textsc{20Newsgroups} is a publicly
  available\footnote{\url{http://qwone.com/~jason/20Newsgroups/}} test
  collection of approximately 20,000 posts on Usenet discussion
  groups, nearly evenly partitioned across 20 different newsgroups.
  In this article we use the ``harder'' version of the dataset, i.e.,
  the one from which all metadata (headers, footers, and quotes) have
  been removed (on this, see also Footnote \ref{foot:harder}).
  % \footnote{This version of the dataset is more challenging than
  % other versions which have been often used in the literature, since
  % headers, footers, and quotes that appear in the documents of
  % \textsc{20Newsgroups} contain information which is extremely
  % revealing of the class the document belongs to.}
  The dataset contains 101,322 distinct terms after removing
  stopwords.
  % \fabscomment{Bisognerebbe almeno dare una citazione a un articolo
  % che la usa.} \alex{L'ho meso nella footnote 9 citando
  % \cite{zhang2015bayesian} (risultati simili ai nostri, cioè soto
  % 0.7, il che vol dire che il reviewer che ci ha segato l'articolo
  % SIGIR usava metadati che hanno una correlazione altissima con i
  % labels -- sopra 0.9)}

\item The \textsc{Ohsumed} test collection \cite{Hersh94} consists of
  a set of
  % 348,566
  MEDLINE documents spanning the years from 1987 to 1991.  Each entry
  consists of summary information relative to a paper published on one
  of 270 medical journals.  The available fields are title, abstract,
  MeSH indexing terms, author, source, and publication type.
  % Not all the entries contain abstract and MeSH
  % indexing terms.
  Following \cite{Joachims98}, we restrict our experiments to the set
  of 23 cardiovascular disease classes, and we use (see
  % , but we adopt the (extended) version discussed at
  \url{http://disi.unitn.it/moschitti/corpora.htm}) the 34,389
  documents of year 1991 that have at least one of these 23 classes.
  Since no standard training/test split has been proposed in the
  literature we randomly partition the set into a part used for
  training (70\% of the documents) and a part used for testing (the
  other 30\%).
  % \fabscomment{Bisognerebbe mettere la citazione a un papero anziché
  % a una URL.} \alex{Non l'ho trovato, pensavo Moschitti l'aveva
  % usato nei suoi articoli ma invece no.}
  The total number of distinct terms after removing stopwords is
  54,949.
  
\end{itemize}

\noindent Since the present work deals with the binary case, each
experiment on each of these test collections here consists of running
as many binary classification tasks as there are classes in the
collection.

% ---------------------------------------------

\subsection{Learning Algorithms}
\label{sec:algorithms}

\noindent As the representation model, in all our experiments we use a
simple unigram model with no stemming or lemmatization. All term
weighting approaches are tested in exactly equal conditions, i.e., we
run each combination of a term weighting method and a learning
algorithm individually for each binary classification problem derived
from each collection.  In all cases, we apply local feature selection
using $\chi^2$ as the feature scoring function and at a reduction
ratio of 0.1; this has proven a good setting in text classification
\cite{Yang97}.

In order to assess the quality of the weighted vectors we consider the
following learning algorithms for training classifiers:
\begin{itemize}
\item Support Vector Machines (SVM) with a linear kernel
  \cite{Fan:2008ys}, an algorithm that finds the hyperplane in
  high-dimensional spaces that separates, by the largest possible
  margin to the nearest training examples, the positive and negative
  examples.
\item Logistic Regression (LR), an algorithm that generates linear
  models of the probability that a document belongs to the class, by
  using the logistic function.  Considering LR in our experiments is
  interesting because the LTW framework we define relies on LR to
  define the weighting function.
  % \footnote{Despite the fact that LR is used as part of the Learning
  % to Weight framework and in the performance evaluation, differences
  % between \bblue{our reimplementation within the framework and the
  % scikit-learn version used as a classifier} \fabscomment{Quali due
  % versioni?} \alex{(modifico il testo)} may
  % exist %, since in the second case we rely on the scikit-learn implementation
  % (see
  % \url{http://scikit-learn.org/stable/modules/linear_model.html\#logistic-regression}
  % for further details).}.
\item Multinomial Naive Bayes (MNB), which implements the na\"ive
  Bayes algorithm for multinomially distributed discrete data (though
  in text classification it is known to work well also for real-valued
  weighted vectors\cite{rennie2003tackling}).
\item K-Nearest Neighbours (KNN), an instance-based learning algorithm
  that outputs the class label more frequent across the $k$ examples
  most similar to the test example; as the measure of similarity we
  use the Euclidean
  distance. %\alex{[Sto ripetendo gli sperimenti KNN per i baselines ampliando il grid-search con feature selection molto agressivo; non fate molto caso ai risultati attuali.]}.
\item Random Forests (RF), a method which builds an ensemble of
  decision trees. The implementation we
  use\footnote{\url{http://scikit-learn.org/stable/modules/generated/sklearn.ensemble.RandomForestClassifier.html}}
  combines the resulting classifiers by averaging their probabilistic
  predictions, instead of returning the most frequent class output by
  the individual trees.
\end{itemize}

\noindent We also include the results of running \textsc{FastText} (FT
-- \cite{Bojanowski:2017fe,Grave:2017wj}), \bblue{a state-of-the-art
method for text classification based on an evolution of the
\textsc{Word2Vec} architecture \cite{Mikolov:2013jk,Mikolov:2013} for
text classification. \textsc{FastText} is not a term weighting method
and is here included for comparison purposes only, i.e., in order to
verify how the text classification pipelines that we use in our
experiments fare with respect to state-of-the-art text classification
methods.  Results for} the \textsc{FastText} classifier are only
reported for the dense representations that \textsc{FastText}
produces, since in its currently available implementation it is not
possible to use the \textsc{FastText} classifier with externally
generated vectors.

% ---------------------------------------------

\subsection{Evaluation Measures}
\label{sec:evaluation}

\noindent As the effectiveness measure we use the well-known $F_{1}$,
the harmonic mean of precision ($\pi$) and recall ($\rho$) defined as
$F_{1}=(2\pi\rho)/(\pi+\rho)=(2TP)/(2TP+FP+FN)$, where $TP$, $FP$,
$FN$, are the numbers of true positives, false positives, false
negatives, from the binary contingency table.  We take $F_{1}=1$ when
$TP=FP=FN=0$, since the classifier has correctly classified all
examples as negative.

In order to average across all the classes of a given dataset we
compute both micro-averaged $F_{1}$ (denoted by $F_{1}^{\mu}$) and
macro-averaged $F_{1}$ (denoted by $F_{1}^M$).  $F_{1}^{\mu}$ is
obtained by (i) computing the class-specific values $TP_{c}$,
$FP_{c}$, and $FN_{c}$, (ii) obtaining $TP$ as the sum of the
$TP_{c}$'s (same for $FP$ and $FN$), and then applying the $F_{1}$
formula.  $F_{1}^M$ is obtained by first computing the class-specific
$F_{1}$ values and then averaging them across the classes.
% \fabscomment{While $F_{1}^M$ attributes equal importance to all
% classes, irrespective of their frequency, frequent classes have a
% higher influence than infrequent ones on$F_{1}^{\mu}$; as a result,
% a classifier that outperforms all others in terms of $F_{1}^M$ is
% necessarily one that handles infrequent classes well, while a
% classifier that outperforms all others in terms of $F_{1}^\mu$ may
% simply be one that performs well on the frequent classes.}  The fact
% that $F_{1}^M$ attributes equal importance to all classes means that
% low-frequency classes will be as important as high-frequency ones in
% determining $F_{1}^M$ scores; $F_{1}^{\mu}$ is instead more
% influenced by high-frequency classes than by low-frequency ones.
% High values of $F_{1}^M$ thus tend to indicate that the classifier
% performs well also on low-prevalence classes, while high values of
% $F_{1}^{\mu}$ may just indicate that the classifier performs well on
% high-prevalence classes.

% ---------------------------------------------

\subsection{Baseline Methods}
\label{sec:baselines}

\noindent We choose some relevant unsupervised and supervised
weighting methods as the baselines against which to compare the
Learning to Weight framework.  The methods considered in this
experimental evaluation are grouped as follows:
\begin{itemize}
\item Unsupervised Term Weighting (UTW) methods: Binary (the simplest
  weighting function, that returns 1 if the document contains the term
  and 0 otherwise), TF (Equation \ref{eq:rawfreq}), LogTF (Equation
  \ref{eq:lognorm}), TFIDF (a variant of Equation \ref{eq:logtfidf}
  that uses the raw frequency of Equation \ref{eq:rawfreq} as the
  $\dd(t,d)$ component),
  % (Equation \ref{eq:tfidf}),
  LogTFIDF (Equation \ref{eq:logtfidf}), and BM25 (Equation
  \ref{eq:bm25}).
\item Supervised Term Weighting (STW) methods:
  % the three pioneer variants TFIG (Equation \ref{eq:ig}),
  TFCHI (Equation \ref{eq:chisquare}) and TFGR (Equation \ref{eq:gr}),
  proposed in \cite{SAC03b}; \emph{ConfWeight} \cite{Soucy:2005fk},
  and TFRF (Equation \ref{eq:rf}), proposed in \cite{Lan:2009kx}.  For
  all STW variants in this evaluation we adopt the $\dd(t,d)$ factor
  defined in Equation \ref{eq:lognorm}. We leave aside TFIG (Equation
  \ref{eq:ig}) since, in binary classification, it is equivalent to
  TFGR, because the only difference between the two is a constant
  normalisation factor that is cancelled out by cosine
  normalisation. We also include the dense representations
  (\emph{Dense}) that \textsc{FastText} produces in this group, since
  they are conditioned on the class labels
  \cite{Bojanowski:2017fe,Grave:2017wj}. \bblue{We also consider the
  CCA method discussed at the end of Section \ref{sec:relatedwork},
  suitably adapted to binary text classification, as a further STW
  baseline.}
\item Learning to Weight (LTW): we consider both local (L -- Equation
  \ref{eq:local}) and global (G -- Equation \ref{eq:global}) versions
  for the $\cd(t,D,C)$ factor. We also investigate a variant whereby
  the sigmoid activation function ($\sigma$) of the $\cd(t,D,C)$
  factor is replaced by an identity function (I -- thus allowing the
  model to output unbounded and negative $\cd(t,D,C)$ scores). We thus
  propose four LTW variants; e.g., LTW-L-$\sigma$ denotes LTW using
  the \underline{L}ocal $\cd(t,D,C)$ variant with the $\sigma$
  activation function.
\end{itemize}

% ---------------------------------------------

\subsection{Implementation Details}
\label{sec:impdetails}

\noindent We have implemented our method using Tensorflow
\cite{abadi2016tensorflow}.  In our experiments we apply dropout to
the hidden layer activations (with a drop probability of 0.2) in order
to prevent overfitting. We use stochastic optimization relying on the
Adam optimiser \cite{adam} with a learning rate of 0.005 (leaving the
rest of the parameters set to their default values, i.e.,
$\beta_1=0.9$, $\beta_2=0.999$, and $\epsilon=1e^{-0.8}$), with a
batch size of 100 and shuffling whenever an epoch is completed.  We
set the maximum number of iterations to 100,000 steps. However, we use
an early stopping criterion, triggered when 20 consecutive validation
steps (each of them run after every 100 training steps) have shown no
improvement. In our experiments the runs obtain this convergence after
an average of 5,780 steps, ranging from 900 to 18,100.  The result
scores that we report for all LTW variants are averages across 10
runs. For the model architecture we use a size of the hidden layer
equal to 1000 for the local variant (which actually is the size of the
convolutional filter that works on each feature separately) and a
larger size of $\frac{1}{2}F$ for the global variant, where $F$ is the
number of distinct terms in the training set. The rationale behind
using a larger size for the global variant is that it is also meant to
model inter-feature relations.
% plot the number of trainable parameters in each case?

For the learning algorithms other than \textsc{FastText} and CCA we
use the implementations provided by Scikit-learn
\cite{scikit-learn}. For \textsc{FastText} we use its publicly
available
implementation\footnote{\url{https://github.com/facebookresearch/fastText}}.
% In its current version, \textsc{FastText} does not take hard
% classification decisions, leaving the number of labels to be
% attributed to a document as a user-defined parameter. This allows
% \textsc{FastText} to be directly applicable to single-label
% scenarios (as is the case of \textsc{20Newsgroups}) by setting this
% parameter to 1. \fabscomment{Discutere con Andrea.}
As with all other learners, also with \textsc{FastText} we train
independent binary classifiers, one for each class.
% In order to apply \textsc{FastText} to our multi-label scenarios
% (\textsc{Reuters-21578} and \textsc{Ohsumed}) we have trained
% independent binary classifiers, one for each class.
We leave the learning rate at 0.1 (its default value). Although the
authors recommend to set the \emph{epoch} parameter (the number of
times a complete pass over the entire set is to be performed) in the
range [5,50], we have experienced a consistent improvement in
performance across all datasets when using higher values. We thus vary
the number of epochs in $\{5,10,25,50,100,250,500\}$, and the best
values turn out to be 100 for \textsc{Reuters-21578} and
\textsc{Ohsumed}, and 500 for \textsc{20Newsgroups} (for which no
further improvement is verified for higher values).

% The same text preprocessing is applied to the documents fed to
% \textsc{FastText} so as to allow for a fair comparison.

\bblue{We have reimplemented\footnote{\bblue{Our implementation of CCA
is accessible as part of our code release.}} the CCA method according
to the specifications in \cite{de2007combined}, and adapted it to
binary classification (in place of ranking, as it was originally
devised for).  This means replacing the ranking-oriented evaluation
function used as the fitness function (originally, a combination of
PAVG and FFP4) with $F_1$, which is better suited for classification
(and is also our evaluation measure of choice). We modify the process
for the selection of the best individual so as to be driven by the
classification accuracy of a logistic regressor\footnote{\bblue{The
choice of logistic regression as a proxy
% for the fitness in CCA is due to two main reasons.  The first
has to do with efficiency. In fact, genetic programming is known to be
computationally expensive, and most of its cost is accounted for by
the evaluation of the fitness function. Since the classifier generated
by logistic regression is an efficient one, this has a positive impact
on the efficiency of this evaluation.
% The second has to do with allowing a fair comparison with our LTW
% approach, since we likewise rely on logistic regression to optimize
% our models.
}} as measured on a validation set.
% (consistently with the optimization strategy we used in our LTW
% variants).
In our implementation of CCA we do not consider terminals $t_{19}$ and
$t_{20}$ since they are inherently defined upon the notion of
``query'', a notion that does not apply to text classification.  All
the hyper-parameters are set to the values recommended in
\cite{de2007combined}.}
 
\bblue{To guarantee fair comparisons between our weighting methods and
the baseline weighting methods, the parameters of each classifier are
optimised individually, i.e., for each $\langle$weighting method,
learning method, binary class$\rangle$ combination. For all such
combinations, optimization is performed on a subset of the training
set used as held-out validation set; once the parameter values have
been chosen, the validation set is merged again into the training set
and the classifier is retrained}.  For SVMs and LR we test values for
the penalty parameter $C$ in the set
$\{ 10^{-4},10^{-3}\ldots 10^{4}\}$, and alternatively consider the
``dual'' and ``primal'' optimization variants. For SVMs we test values
for the \emph{loss} parameter in $\{hinge,hinge^2\}$. Adhering to a
practice well documented in the literature, for SVMs we adopt the RBF
kernel in the experiments in which dense vectors are used (i.e., in
the experiments that use the representations produced by
\textsc{FastText}, documented in
% , which have 100 dimensions
rows ``Dense'' of Tables \ref{tab:macrof1} and \ref{tab:microf1}),
while we adopt the linear kernel when documents are represented by
sparse vectors (i.e., all other experiments).  For LR we test both L1
and L2 regularisation. For MNB, for the $\alpha$ parameter we test all
values in \{0.0, 0.001, 0.01, 0.05, 0.1, 1.0\}.  For KNN, for the $k$
parameter we test all values in $\{1, 3, 5, 15, 30\}$ (a) with all
features, or (b) with dimensionality reduction obtained by selecting
the top $\{25, 50, 100, 250, 500\}$ features using the $\chi^2$
feature scoring function, or (c) with dimensionality reduction
obtained via Principal Component Analysis (PCA) at $\{64,128,256\}$
dimensions.  For RF we vary the number of trees in
$n\in\{10,25,50,100\}$, we test both the \emph{Gini} and
\emph{Entropy} criteria, and we consider a maximum number of features
in $\{\sqrt{F}, \log{F}, 1000\}$, where $F$ is the total number of
features.

% ---------------------------------------------

\subsection{Results}
\label{sec:results}

\noindent Tables \ref{tab:macrof1} and \ref{tab:microf1} report the
results we have obtained on the \textsc{Reuters-21578},
\textsc{20Newsgroups}\footnote{\label{foot:harder}While some previous
papers (e.g., \cite{salles2015broof}) have reported substantially
higher scores for this dataset, it is worth noticing that we use a
harder, more realistic version of the dataset than has been used in
those papers.  In our version, headers, footers, and quotes have been
removed, since these fields contain terms that have near-perfect
correlation with the classes of interest, thus making the
classification task unrealistically easy; see
\url{http://scikit-learn.org/stable/datasets/twenty_newsgroups.html}
for further details. Our results are indeed consistent with those of
other papers (e.g., \cite{zhang2015bayesian}) which follow the same
policy as ours.}, and \textsc{Ohsumed} datasets, for macro- and
micro-averaged $F_{1}$.  Values in boldface indicate the best results
obtained with the given learner on the given dataset, while values in
greyed-out cells indicate the LTW variants that outperform all
baselines.
%
% -----------------------------------------------------------------------
% Macro F1
% -----------------------------------------------------------------------
\begin{table*}[ht!]
  \centering
  \resizebox{\textwidth}{!}{%
  \begin{tabular}{|l|l||c|c|c|c|c|c||c|c|c|c|c|c||c|c|c|c|c|c|}
    \hline
    & \textbf{} & \multicolumn{6}{c||}{\textsc{Reuters-21578}} & \multicolumn{6}{c||}{\textsc{20Newsgroups}} & \multicolumn{6}{c|}{\textsc{Ohsumed}} \\ \hline\hline
    \textbf{} & \textbf{$F^{M}_{1}$} & \textbf{SVM} & \textbf{LR} & \textbf{MNB} & \textbf{KNN} & \textbf{RF} & \textbf{FT} & \textbf{SVM} & \textbf{LR} & \textbf{MNB} & \textbf{KNN} & \textbf{RF} & \textbf{FT}  & \textbf{SVM} & \textbf{LR} & \textbf{MNB} & \textbf{KNN} & \textbf{RF} & \textbf{FT} \\ \hline\hline
    \multirow{6}{*}{\begin{sideways}UTW\end{sideways}} & \textbf{Binary} & .519 & .579 & .489 & .493 & .562 & -- & .581 & .593 & .615 & .429 & .579 & -- & .578 & .601 & .548 & .516 & .626 & -- \\ \cline{2-20} 
    & \textbf{TF} & .592 & .599 & .490 & .496 & .555 & -- & .583 & .579 & .601 & .418 & .578 & -- & .593 & .606 & .540 & .530 & .634 & -- \\ \cline{2-20} 
    & \textbf{LogTF} & .569 & .612 & .449 & .522 & .558 & -- & .611 & .619 & .571 & .470 & .570 & -- & .620 & .632 & .453 & .550 & .638 & -- \\ \cline{2-20} 
    & \textbf{TFIDF} & .596 & .604 & .473 & .502 & .544 & -- & .598 & .609 & .594 & .492 & \textbf{.592} & -- & .605 & .612 & .506 & .558 & .639 & -- \\ \cline{2-20} 
    & \textbf{LogTFIDF} & .576 & .609 & .462 & .496 & .530 & -- & .607 & .622 & .580 & .512 & .583 & -- & .614 & .636 & .491 & .574 & .643 & -- \\ \cline{2-20} 
    & \textbf{BM25} & .554 & .582 & .462 & .509 & .561 & -- & .602 & .621 & .572 & .536 & .588 & -- & .607 & .634 & .473 & .576 & .637 & -- \\ \cline{2-20} 
    \hline\hline
    
    \multirow{6}{*}{\begin{sideways}STW\end{sideways}}
    & \textbf{TFGR} & .598 & \textbf{.623} & \textbf{.539} & .561 & .572 & -- & .613 & .613 & .653 & .500 & .571 & -- & .508 & .546 & .299 & .566 & .621 & -- \\ \cline{2-20} 
    & \textbf{TFCHI} & .591 & .608 & .514 & .538 & \textbf{.579} & -- & .585 & .590 & .645 & .509 & .587 & -- & .478 & .514 & .267 & .590 & .626 & -- \\ \cline{2-20} 
    & \textbf{ConfWeight} &  .580 &  .587 &  .497 &  .543 &  .540 & -- &  .588 &  .577 &  .578 &  .485 &  .569 & -- &  .649 &  .647 &  \textbf{.617} &  .599 &  \textbf{.646} & -- \\ \cline{2-20} 
    & \textbf{TFRF} & .584 & .602 & .461 & .508 & .539 & -- & .626 & .624 & .617 & .434 & .581 & -- & .635 & .634 & .552 & .477 & .640 & -- \\ \cline{2-20} 
    % & \textbf{\revised{Dense}} & \revised{.541} & \revised{.537} &
    % \revised{.508} & \revised{.565} & \revised{.547} &
    % \revised{.553} & \revised{.616} & \revised{.610} &
    % \revised{.604} & \revised{\textbf{.620}} &
    % \revised{\textbf{.613}} & \revised{.623} & \revised{.618} &
    % \revised{.619} & \revised{.533} & \revised{.620} &
    % \revised{.617} & \revised{.617} \\ \hline \hline
    & \textbf{Dense} & .541 & .537 & .508 & .565 & .547 & .553 & .574 & .572 & .539 & .582 & .585 & .566 & .618 & .619 & .533 & .620 & .617 & .617  \\ \cline{2-20} 

    & \textbf{\bblue{CCA}} & .\bblue{556} & .\bblue{574} & .\bblue{418} & .\bblue{511} & .\bblue{562} & -- & .\bblue{581} & .\bblue{578} & .\bblue{604} & .\bblue{487} & .\bblue{579} & -- & .\bblue{616} & .\bblue{612} & .\bblue{572} & .\bblue{561} & .\bblue{635} & -- \\ \hline \hline  

    \multirow{4}{*}{\begin{sideways}LTW\end{sideways}}
    & \textbf{LTW-L-$\sigma$} & \shadow.614 & .610 & .517 & .542 & .547 & -- & .625 & .619 & \shadow.660 & \shadow\textbf{.600} & .577 & -- & .642 & .642 & .596 & .543 & .624 & -- \\ \cline{2-20} 
    & \textbf{LTW-L-I} & \shadow\textbf{.629} & .619 & .502 & \shadow\textbf{.583} & .553 & -- & \shadow.630 & \shadow\textbf{.626} & \shadow\textbf{.668} & .582 & .573 & -- & .649 & .645 & .607 & .576 & .616 & -- \\ \cline{2-20} 
    & \textbf{LTW-G-$\sigma$} & .531 & .530 & .375 & .442 & .505 & -- & .541 & .534 & .537 & .278 & .514 & -- & .554 & .548 & .480 & .343 & .566 & -- \\ \cline{2-20} 
    & \textbf{LTW-G-I} & \shadow.604 & .604 & .374 & .555 & .540 & -- & \shadow\textbf{.635} & .620 & .494 & 573 & .571 & -- & \shadow\textbf{.651} & \shadow\textbf{.650} & .459 & \shadow\textbf{.638} & .633 & -- \\ \hline 
  \end{tabular}%
  }
  \caption{Results on \textsc{Reuters-21578}, \textsc{20Newsgroups} and \textsc{Ohsumed} in terms of $F^{M}_{1}$.}
  \label{tab:macrof1}
\end{table*}

% -----------------------------------------------------------------------
% micro F1
% -----------------------------------------------------------------------

\begin{table*}[ht!]
  \centering
  \resizebox{\textwidth}{!}{%
  \begin{tabular}{|l|l||c|c|c|c|c|c||c|c|c|c|c|c||c|c|c|c|c|c|}
    \hline
    & \textbf{} & \multicolumn{6}{c||}{\textsc{Reuters-21578}} & \multicolumn{6}{c||}{\textsc{20Newsgroups}} & \multicolumn{6}{c|}{\textsc{Ohsumed}} \\ \hline\hline
    \textbf{} & \textbf{$F^{\mu}_{1}$} & \textbf{SVM} & \textbf{LR} & \textbf{MNB} & \textbf{KNN} & \textbf{RF} & \textbf{FT} & \textbf{SVM} & \textbf{LR} & \textbf{MNB} & \textbf{KNN} & \textbf{RF} & \textbf{FT} & \textbf{SVM} & \textbf{LR} & \textbf{MNB} & \textbf{KNN} & \textbf{RF} & \textbf{FT} \\ \hline\hline
    \multirow{6}{*}{\begin{sideways}UTW\end{sideways}} & \textbf{Binary} & .822 & .843 & .649 & .794 & .841 & -- & .595 & .609 & .623 & .449 & .595 & -- & .642 & .639 & .589 & .547 & .663 & -- \\ \cline{2-20} 
    & \textbf{TF} & .836 & .843 & .621 & .797 & .847 & -- & .596 & .592 & .598 & .436 & .593 & -- & .643 & .643 & .583 & .559 & .684 & -- \\ \cline{2-20} 
    & \textbf{LogTF} & .855 & .861 & .786 & .815 & .846 & -- & .628 & .634 & .599 & .492 & .593 & -- & .663 & .668 & .550 & .580 & .684 & -- \\ \cline{2-20} 
    & \textbf{TFIDF} & .852 & .850 & .787 & .818 & .839 & -- & .617 & .627 & .619 & .511 & \textbf{.612} & -- & .659 & .612 & .574 & .581 & .681 & -- \\ \cline{2-20} 
    & \textbf{LogTFIDF} & .849 & .861 & .781 & .816 & .841 & -- & .627 & .637 & .607 & .532 & .605 & -- & .665 & .668 & .566 & .599 & .684 & -- \\ \cline{2-20} 
    & \textbf{BM25} & .844 & .850 & .769 & .819 & .847 & -- & .622 & .638 & .599 & .552 & .610 & -- & .652 & .659 & .549 & .604 & \textbf{.685} & -- \\ \cline{2-20} 
    \hline\hline
    \multirow{6}{*}{\begin{sideways}STW\end{sideways}}
    & \textbf{TFGR} & .846 & .854 & .821 & .815 & \textbf{.849} & -- & .630 & .631 & .667 & .516 & .589 & -- & .577 & .587 & .410 & .596 & .664 & -- \\ \cline{2-20} 
    & \textbf{TFCHI} & .836 & .839 & .798 & .805 & .846 & -- & .601 & .607 & .659 & .516 & .602 & -- & .566 & .588 & .379 & .616 & .665 & -- \\ \cline{2-20} 
    & \textbf{ConfWeight} & .823 & .821 & .754 & .808 & .834 & -- & .603 & .592 & .609 & .495 & .593 & -- & .668 & .677 & .657 & .635 & .682 & -- \\ \cline{2-20} 
    & \textbf{TFRF} & .862 & .864 & .790 & .804 & .840 & -- & .643 & .637 & .644 & .452 & .602 & -- & .677 & .662 & .619 & .536 & .679 & -- \\ \cline{2-20} 
    % & \textbf{\revised{Dense}} & \revised{.849} & \revised{.851} &
    % \revised{.821} & \revised{\textbf{.851}} & \revised{.848} &
    % \revised{.851} & \revised{.623} & \revised{.618} &
    % \revised{.619} & \revised{\textbf{.628}} &
    % \revised{\textbf{.621}} & \revised{.631} & \revised{.647} &
    % \revised{.648} & \revised{.558} & \revised{.548} &
    % \revised{.646} & \revised{.651} \\ \hline \hline
    & \textbf{Dense} & .849 & .851 & .821 & \textbf{.851} & .848 & .851 & .580 & .579 & .551 & .591 & .597 & .575 & .647 & .648 & .558 & .548 & .646 & .651 \\ \cline{2-20} 
    & \textbf{CCA} & .\bblue{837} & .\bblue{841} & .\bblue{623} & .\bblue{406} & .\bblue{838} & -- & .\bblue{600} & .\bblue{601} & .\bblue{605} & .\bblue{507} & .\bblue{597} & -- & .\bblue{672} & .\bblue{673} & .\bblue{615} & .\bblue{595} & .\bblue{685} & -- \\ \hline \hline 
    \multirow{4}{*}{\begin{sideways}LTW\end{sideways}}
    & \textbf{LTW-L-$\sigma$} & \shadow.867 & \shadow.865 & \shadow\textbf{.825} & .819 & .848  & -- & \shadow.645 & .636 & \shadow.685 & \shadow\textbf{.603} & .597 & -- & \shadow.688 & \shadow.678 & \shadow.660 & .598 & .663 & -- \\ \cline{2-20} 
    & \textbf{LTW-L-I} & \shadow\textbf{.874} & \shadow\textbf{.869} & .821 & .843 & .846 & -- & \shadow.649 & \shadow\textbf{.643} & \shadow\textbf{.690} & .586 & .593 & -- &  \shadow.688 & \shadow.680 & \shadow\textbf{.665} & .622 & .650 & -- \\ \cline{2-20} 
    & \textbf{LTW-G-$\sigma$} & .824 & .818 & .773 & .776 & .819 & -- & .555 & .546 & .563 & .292 & .536 & -- & .601 & .595 & .555 & .419 & .615 & -- \\ \cline{2-20} 
    & \textbf{LTW-G-I} & \shadow.869 & \shadow.867 & .699 & 830 & .844 & -- & \shadow\textbf{.650} & .635 & .516 & .589 & .590 & -- & \shadow\textbf{.693} & \shadow\textbf{.689} & .513 & \shadow\textbf{.666} & .663 & -- \\ \hline 
  \end{tabular}%
  }
  \caption{Results on \textsc{Reuters-21578}, \textsc{20Newsgroups} and \textsc{Ohsumed} in terms of $F^{\mu}_{1}$.}
  \label{tab:microf1}
\end{table*}

The results indicate that most LTW approaches perform comparably or
better than the baselines in terms of $F^{M}_{1}$, and outperform the
baselines in $F^{\mu}_{1}$ in most cases (statistical significance is
discussed in Section \ref{sub:stat}). The best absolute classification
result for each dataset is always obtained by LTW.

The local variants exhibit the most stable improvement across datasets
and classifiers, while the global variants seem more unstable in this
regard.  The local variants outperform all other baselines on average
in \textsc{Reuters-21578} and \textsc{20Newsgroups} (and rank second
in \textsc{Ohsumed}); the LTW-L-I variant is overall the
best-performing method. The LTW-G-I vectors produce competitive
results when used in combination with SVMs, LR, and KNN.  Conversely,
the globally computed vectors prove less suitable for the MNB
classifier; this may be explained by the fact that globally leveraging
the dependencies between features contradicts the independence
assumption built into the MNB classifier.

It is worth noticing that the unbounded versions (I) turn out to be
more competitive than the bounded ones ($\sigma$).  This seems to
clash with the fact that traditional implementations of $\cd(t,D,C)$
generate values that are always non-negative and usually upper-bounded
by 1.  However, allowing the score to be negative gives the optimiser
the opportunity to discern between positive and negative correlation
with the class, while most FS functions do not draw any distinction
between these two cases.  Furthermore, allowing the score to exceed
the bounds [0,1] helps the optimiser to tune the relative importance
of the $\dd(t,d)$ and the $\cd(t,D,C)$
factors.%, e.g., returning higher values for the $\cd(t,D,C)$ factor
% will diminish the importance of the tf.

Regarding the baselines, TFGR and TFCHI outperform the UTW approaches
in most cases across \textsc{Reuters-21578} and \textsc{20Newsgroups},
but perform comparably worse in \textsc{Ohsumed}. This seems to
indicate that, although traditional FS functions succeed in reflecting
feature importance, some adaptation may be required in order to safely
incorporate this score into the weight calculation.
% We found no consistent differences in performance between the TFIG
% and TFGR, which we deem to the fact that term weighting is operating
% at the binary level (i.e., for each distinct class), and therefore
% the normalisation factor of GR has low impact in the
% calculation. \alex{Infatti ho capito dopo che le variazioni nei
% risultati erano dovute all'aleatorietà dell'algoritmo di
% ottimizazione. Ho tolto TFIG.}
Binary is the worst-performing method (even though it works reasonably
well with the MNB classifier -- which was to be expected given that
the classifier
% , despite being able to deal with real-valued vectors
% \cite{rennie2003tackling},
was originally designed with binary features in mind\footnote{The
particular implementation we use here is able to take advantage of
real-valued vectors, though. See \cite{rennie2003tackling} for further
details.}).
% \fabscomment{Non capisco quest'ultima frase: MNB dovrebbe funzionare
% SOLO con binary features ...} \alex{Questa implementazione usa pure
% i valori pesati.}
This is a consequence of the fact that binary weighting disregards
term frequency (apart from mere presence/absence) and term specificity
in the collection.  Other things being equal, it is noteworthy that
both UTW and STW perform irregularly across different conditions
(e.g., in the STW group \emph{ConfWeight} obtained the best average
performance on \textsc{Ohsumed} but the worst on
\textsc{Reuters-21578} and \textsc{20Newsgroups}), without any clear
winner in the respective groups.
% some commentaries about macro and micro results?  some commentaries
% about the logistic regression (in the sense that it is more related
% to our approach)

The dense representations generated by \textsc{FastText} perform
% well in combination with KNN and RF classifiers, as witnessed by the
% fact that they outperform all other competitors in
% \textsc{20Newsgroups} and perform always comparably to the
% best-performing competitor in the other cases.
comparably to the best-performing competitor in combination with KNN
and RF classifiers in most cases.  As a classifier, \textsc{FastText}
(FT) performed irregularly across datasets, outperforming several
baselines on \textsc{Ohsumed}, both in terms of $F_1^M$ and
$F_1^{\mu}$, but
% beating all UTW-equipped and most STW-equipped classifiers (with the
% exception of those using TRFR and, in some cases, TFGR and TFCHI) in
% \textsc{20Newsgroups}, but
being surpassed by many UTW and STW methods in combination with SVM
and LR on the other datasets.  Those results seem to confirm that the
superiority of deep learning models over traditional learners is only
reached
% brought forward
when learning from much larger text collections.
% \textsc{Reuters-21578} and \textsc{Ohsumed}. Not in vain, these last
% two collections are multi-label and, as recalled from Section
% \ref{sec:impdetails}, \textsc{FastText} is better fit to operate in
% single-label domains.} \fabscomment{Questa ultima frase è da
% rivedere in base ai nuovi esperimenti.

\bblue{In our experiments CCA never stands out from the point of view
of accuracy, obtaining scores which are almost always in-between the
worst result and the best result obtained by the other
methods. Despite the fact that CCA is somehow similar in spirit to
LTW, in the sense that both frameworks aim at optimizing the term
weighting function (following radically different strategies, though),
the weighting functions produced by CCA are often too complex to
interpret.  In our experiments, the depth of the formulae (when viewed
as trees of operators and operands) representing the weighting
functions that CCA finds, range from a minimum of 1 to a maximum of
13; the mode of the distribution is 5, which already indicates fairly
complex formulae.
% In LTW approaches, on the contrary, the depth is always 2.
%% has been designed to facilitate a subsequent inspection of the
%% weighting function, akin to the plots produced in ROC analysis.
}
% \fabscomment{Quest'ultima frase non è superconvincente.}

%
% \begin{figure}[h!]
%   \centering
%   \includegraphics[height=22cm]{Figures/plots_pos}
%   \caption{Plots of LTW-L-$\sigma$ variants. From top to bottom
%   \fabscomment{top to bottom}: \textsc{Reuters-21578} (earn),
%   \textsc{20Newsgroups} (misc.forsale), \textsc{Ohsumed}
%   (Respiratory Tract Diseases).  }
%   \label{fig:norm_pos}
% \end{figure}
%%
% \begin{figure}[h!]
%   \centering
%   \includegraphics[height=22cm]{Figures/plots_posneg}
%   \caption{Plots of LTW-L-I variants. From top to bottom
%   \fabscomment{top to bottom}: \textsc{Reuters-21578} (earn),
%   \textsc{20Newsgroups} (misc.forsale), \textsc{Ohsumed}
%   (Respiratory Tract Diseases).  }
%   \label{fig:norm_posneg}
% \end{figure}

\begin{figure*}[h!]
  \centering
  \begin{subfigure}[b]{.9\columnwidth}
    \includegraphics[height=22cm,center]{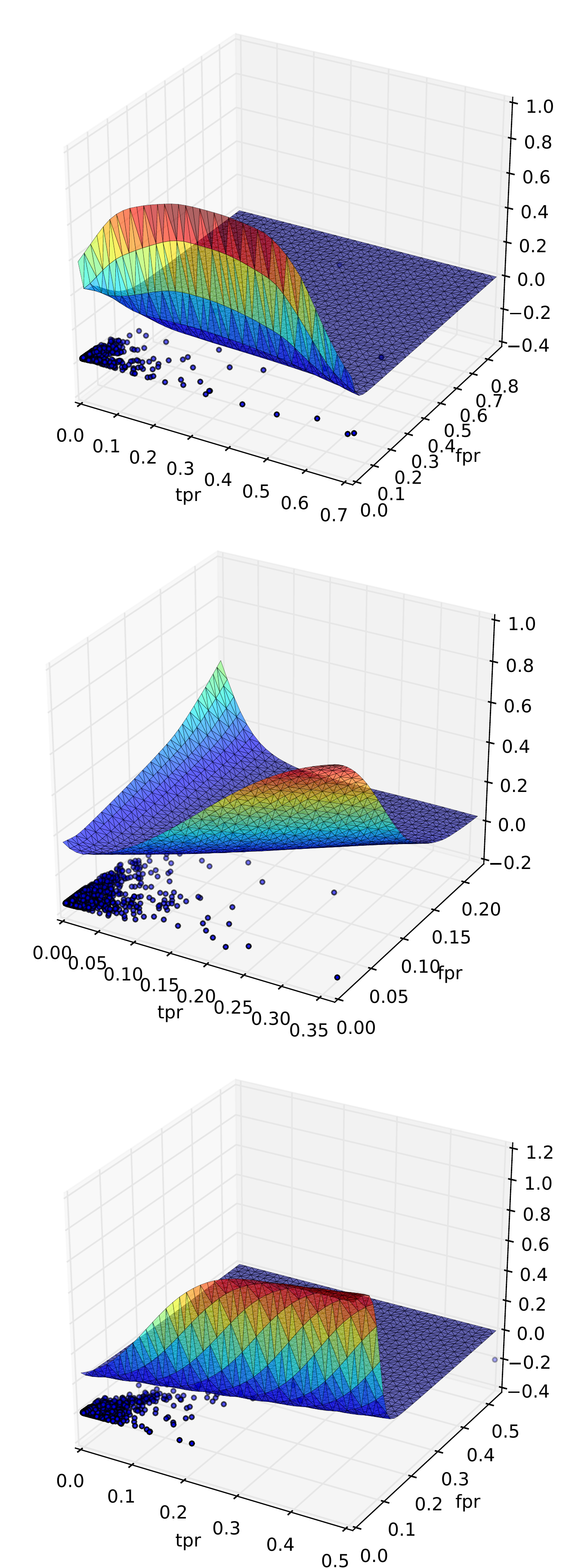}
    \caption{}\label{fig:norm_pos}
  \end{subfigure} \quad
  \begin{subfigure}[b]{\columnwidth}
    \includegraphics[height=22cm,center]{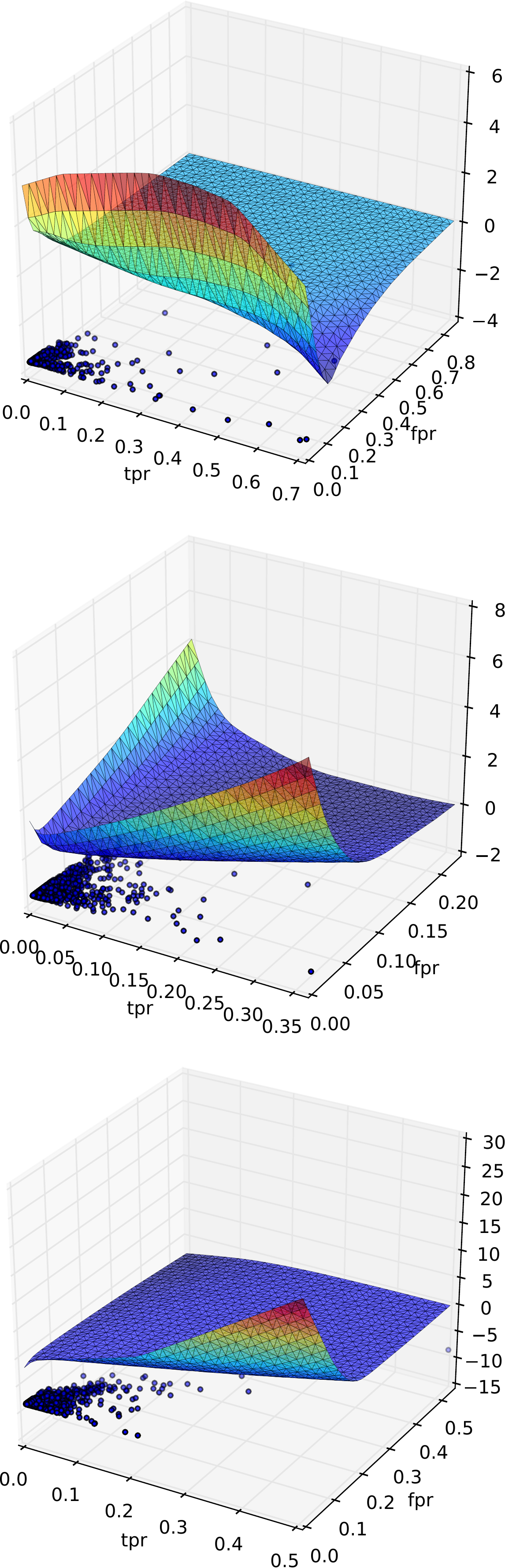}
    \caption{}\label{fig:norm_posneg}
  \end{subfigure}
  \caption{Plots of the LTW-L-$\sigma$ variant (a) and LTW-L-I variant
  (b) on (from top to bottom) \textsc{Reuters-21578} (earn),
  \textsc{20Newsgroups} (misc.forsale), \textsc{Ohsumed} (Respiratory
  Tract Diseases).}
  \label{fig:plotsofLTW-L}
\end{figure*}

Figures \ref{fig:norm_pos} and \ref{fig:norm_posneg} show examples of
the learned local $\cd(t,D,C)$ functions in their LTW-L-$\sigma$ and
LTW-L-I variants, respectively, for a sample class in each dataset. We
also include the actual coordinates ($\tpr$,$\fpr$) for each of the
features. (Similar plots cannot be displayed for the global variants
since global $\cd(t,D,C)$ functions depend on a larger number of
coordinates.)  Note that all learned functions agree that the most
important terms are those characterized by a high $\tpr$ and a very
low $\fpr$.  However, different cases give rise to different
shapes\footnote{We verified that the shapes are consistent through
different runs, with negligible variations among them. In additional
experiments we forced the $\cd(t,D,C)$ function to mimic information
gain in a pre-training phase. This had essentially no impact on the
final
shape.}. % which help to produce more linearly separable examples during optimization.
The LTW-L-I versions tends to generate values that are sometimes
higher than 1, and even smaller than 0 under certain conditions, e.g.,
when $\fpr$ moves away from low values (see Figure
\ref{fig:norm_posneg}, top) or when $\tpr$ has very low values (see
Figure \ref{fig:norm_posneg}, bottom).  Note also that the plots from
\textsc{20Newsgroups} bear some resemblance to information gain (see
Figure \ref{fig:fs_as_stw}, top), while the plots from
\textsc{Ohsumed} behave somehow similarly to the ones relative to
relevance frequency (Figure \ref{fig:chi_gr_rf}, bottom).  Besides the
fact that there exists some fundamental differences between the two
(e.g., the one resembling information gain is almost symmetric with
respect to the polarity of the correlation, while the other is not),
it is interesting to see how LTW ``decides'' automatically the best
shape to develop for each of the binary problems at hand.  That this
function is potentially different for each problem was not necessarily
to be taken for granted, i.e., the optimised function might have
exhibited a similar pattern across problems.  These examples support
our intuition that the optimal $\cd(t,D,C)$ function may neither be
unique, nor universal (thus, not liable to be captured by a fixed
formula, as done in standard weighting approaches, supervised and
unsupervised alike), but can instead be learned for each specific
classification problem.

% ---------------------------------------------

\subsection{Statistical Significance}\label{sub:stat}

% \noindent We found differences in $F^{M}_{1}$ between the local
% variants and the baselines to be statistically significant according
% to a two-tailed paired t-test at a cutoff of $\alpha=0.05$ (and
% $\alpha=0.005$ in most cases), with the exception of TFGR and TFCHI,
% which are not significantly different from the local versions, and
% ConfWeight, which is not significantly different from
% LTW-L-$\sigma$.  In terms of $F^{\mu}_{1}$, differences in
% performance for both local variants were found to be statistically
% significant with respect to all baselines at $\alpha=0.05$ (and at
% $\alpha=0.005$ in most cases), see Table \ref{tab:ttest}.
\noindent In Table \ref{tab:ttest} we report the results of our
statistical significance tests. The differences between the local
variants of LTW and the UTW baselines are statistically significant in
all cases. %, both in terms of $F^{M}_{1}$ and $F^{\mu}_{1}$.
In terms of $F^{M}_{1}$ and regarding the STW approaches,
LTW-L-$\sigma$ is significantly superior only to TFRF, while LTW-L-I
significantly outperforms all STW baselines but TFGR and Dense.  In
terms of $F^{\mu}_{1}$, both LTW-L-$\sigma$ and LTW-L-I are always
superior to the STW baselines.  There are no significant differences
in performance between the local variants according to the test.
Finally, the global variants do not prove superior, in a statistically
significance sense, to any of the baselines. Concerning the global
variants, LTW-G-I always proves superior to LTW-G-$\sigma$.

\begin{table}
  \resizebox{\columnwidth}{!}{
  \begin{tabular}{|c|l||c|c|c|c|c|c||c|c|c|c|c|c||c|c|c|c|}
    \hline
    & & \multicolumn{6}{c||}{UTW} & \multicolumn{6}{c||}{STW} & \multicolumn{4}{c|}{LTW} 
    \\ 
    \hline
    & & \begin{sideways}Binary\end{sideways} & \begin{sideways}TF\end{sideways} & \begin{sideways}LogTF\end{sideways} & \begin{sideways}TFIDF\end{sideways} & \begin{sideways}LogTFIDF\end{sideways} & \begin{sideways}BM25\end{sideways} & \begin{sideways}TFGR\end{sideways} & \begin{sideways}TFCHI\end{sideways} & \begin{sideways}ConfWeight\end{sideways} & \begin{sideways}TFRF\end{sideways} & \begin{sideways}Dense\end{sideways} & \begin{sideways}\bblue{CCA}\end{sideways} & \begin{sideways}LW-L-$\sigma$\end{sideways} & \begin{sideways}LW-L-I\end{sideways} & \begin{sideways}LW-G-$\sigma$\end{sideways} & \begin{sideways}LW-G-I\end{sideways} \\\hline\hline
    \multirow{4}{*}{\begin{sideways}$F_1^M$\end{sideways}} 
    & LW-L-$\sigma$ & $\dag\dag$ & $\dag\dag$ & $\dag$ & $\dag$ & $\dag$ & $\dag$ &   &   &   & $\dag$ & $\dag$ & \bblue{$\dag\dag$} &  -  &   & $\dag\dag$ &   \\\cline{2-18}
    & LW-L-I & $\dag\dag$ & $\dag\dag$ & $\dag\dag$ & $\dag$ & $\dag$ & $\dag$ & $\dag$ & $\dag$ & $\dag$ & $\dag$ & $\dag$ & \bblue{$\dag\dag$} &   &  -  & $\dag\dag$ &   \\\cline{2-18}
    & LW-G-$\sigma$ &   &   &   &   &   &   &   &   &   &   &   &   &   &   &  -  &   \\\cline{2-18}
    & LW-G-I &   &   &   &   &   &   &   &   &   &   &   &   &   &   & $\dag\dag$ &  -  \\\hline\hline
    \multirow{4}{*}{\begin{sideways}$F_1^{\mu}$\end{sideways}} 
    & LW-L-$\sigma$ & $\dag\dag$ & $\dag\dag$ & $\dag\dag$ & $\dag$ & $\dag$ & $\dag$ & $\dag\dag$ & $\dag\dag$ & $\dag$ & $\dag$ & $\dag$ & \bblue{$\dag\dag$} &  -  &   & $\dag\dag$ &   \\\cline{2-18}
    & LW-L-I & $\dag\dag$ & $\dag\dag$ & $\dag$ & $\dag$ & $\dag$ & $\dag$ & $\dag\dag$ & $\dag\dag$ & $\dag$ & $\dag$ & $\dag$ & \bblue{$\dag\dag$} &   &  -  & $\dag\dag$ &   \\\cline{2-18}
    & LW-G-$\sigma$ &   &   &   &   &   &   &   &   &   &   &   &   &   &   &  -  &   \\\cline{2-18}
    & LW-G-I &   &   &   &   &   &   &   &   &   &   &   &   &   &   & $\dag$ &  -  \\\hline
  \end{tabular}
  }
  \caption{Wilcoxon signed-rank tests of statistical significance of the difference in performance between LTW variants (rows) and all tested methods (columns), at confidence levels $\alpha=0.05$ ($\dag$) and $\alpha=0.005$ ($\dag\dag$).}
  \label{tab:ttest}
\end{table}

Since the optimization procedure has a random component we analyze the
variation in performance across the 10 runs in terms of standard
deviation (SD); no difference worth noticing results from different
random seed initializations.  Specifically, the SD of $F^{M}_{1}$
across runs varies between $0.0018$ (in \textsc{20Newsgroups} using LR
and LTW-L-I) and $0.0178$ (in \textsc{Reuters-21578} using MNB and
LTW-L-$\sigma$), with an expected value of $0.0082$.  Similarly, the
SD of $F^{\mu}_{1}$ varies between $0.0017$ (in \textsc{20Newsgroups}
using SVM and LTW-L-I) and $0.0179$ (in \textsc{Ohsumed} using MNB and
LTW-G-I), with an expected value of $0.0072$.  All in all, we find the
local variants to be slightly more stable (in terms of SD) than the
global ones across learners and datasets ($0.0076$ vs. $0.0088$ in
$F^{M}_{1}$, and $0.0043$ vs. $0.0099$ in $F^{\mu}_{1}$,
respectively), while there are no noticeable differences between
bounded and unbounded versions in this respect.

% ---------------------------------------------

\subsection{A Note on Convergence and
Efficiency}\label{sub:efficiency}

\begin{figure}[ht!]
  % \centering
  % \includegraphics[width=\textwidth]{Figures/convergence.png}
  \includegraphics[width=\columnwidth]{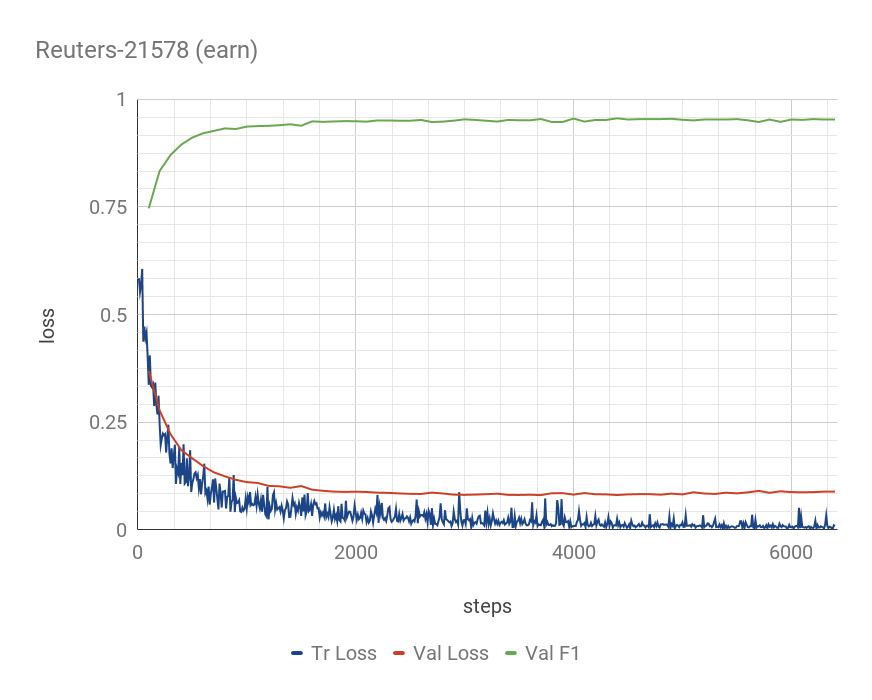}
  \includegraphics[width=\columnwidth]{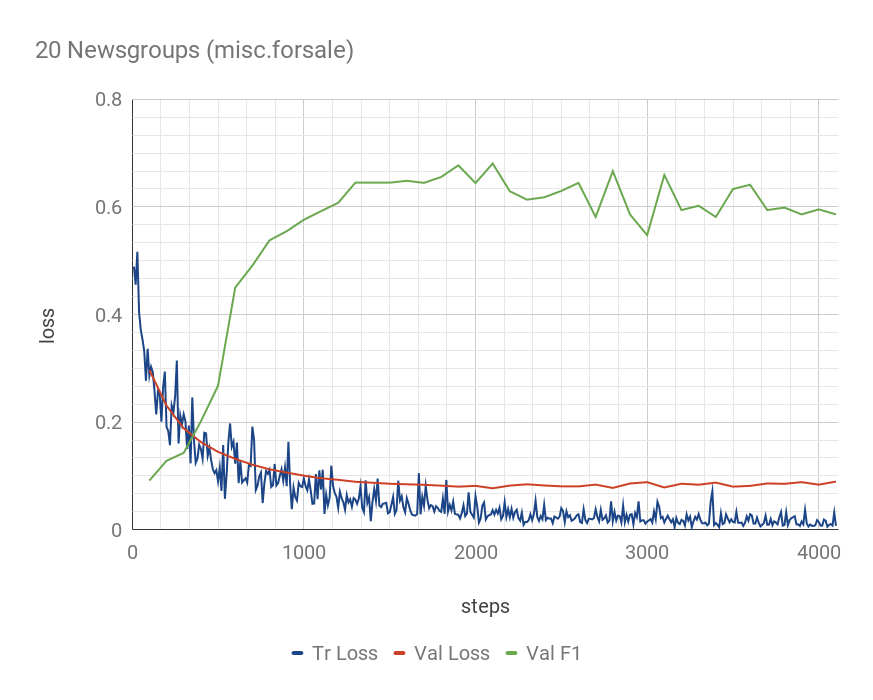}
  \includegraphics[width=\columnwidth]{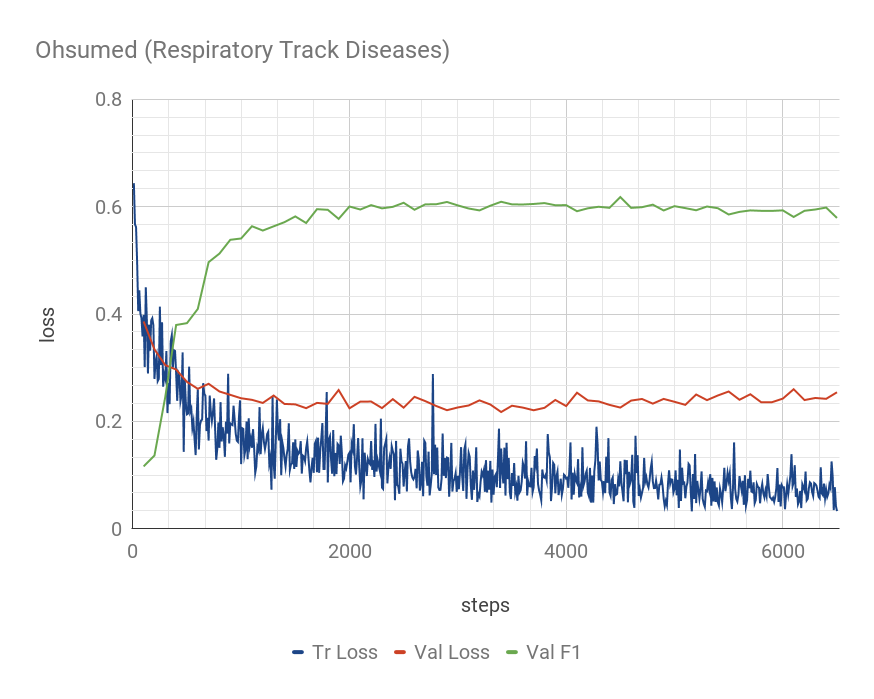}
  \vspace{0.3cm}
  \caption{Convergence trends of LTW-L-I variants. From top to bottom:
  \textsc{Reuters-21578} (earn), \textsc{20Newsgroups} (misc.forsale),
  \textsc{Ohsumed} (Respiratory Tract Diseases). }
  \label{fig:convergence}
\end{figure}

\noindent As a learning procedure, LTW is exposed to the typical
problems that arise in the realm of optimization methods.
Notwithstanding this, we observe that the different models converge
smoothly to good solutions in the parameter space (as demonstrated in
classification performance), with low tendency to overfit. Figure
\ref{fig:convergence} shows the convergence trends for LTW-L-I for the
same sample classes used in Figure \ref{fig:plotsofLTW-L} (we observe
that all the LTW variants we propose and all the classes from our
three datasets exhibit a qualitatively similar behaviour in terms of
convergence). Note that, in all cases, an early termination is
activated before the maximum number of iterations is reached. That is,
irrespectively of the ongoing reduction in training loss, the process
terminates when no further improvement is recorded in a validation
set. Apart from preventing overfitting, this mechanism speeds up the
optimization time noticeably.

Although execution times depend largely on the implementation and the
hardware, it is fair to note that LTW approaches are slower than UTW
methods and most STW methods \bblue{(with the exception of CCA)}.  For
example, on the same machine\footnote{\bblue{All timings were recorded
on the same machine, equipped with a 12-core processor Intel Core
i7-4930K at 3.40GHz with 32 GB of RAM and an Nvidia GeForce GTX 1080,
under Ubuntu 16.04 (LTS)}} and on average, it tales less than a second
for any UTW to compute term weights for the \textsc{Reuters-21578}
collection, while STW requires 23 seconds and LTW requires 62
seconds. Despite the increase in execution time, it is also fair to
note that this penalty is affordable, since the time bottleneck still
lies on the tasks of training the classifier and optimizing the
hyperparameters in the validation phase, which are run once for
all. Additionally, given the continuous improvements and massive
parallelization of new GPUs, it would not be surprising to see this
difference in performance to considerably soften in the near
future.\footnote{\textsc{FastText} is peculiar in this respect, since
the document representation phase and the classifier training phase
are undertaken simultaneously. On average, it tales 10 seconds to
process \textsc{Reuters-21578}, 31 seconds to process \textsc{Ohsumed}
(we use 100 epochs for both), and 79 seconds to process
\textsc{20Newsgroups} (500 epochs).}
% \fabscomment{Considerare l'ipotesi di spostare la footnote più in
% basso.}
\bblue{CCA is by far the slowest method among those considered in our
experiments: although the execution time in any genetic algorithm
depends on many variables (number of iterations, population size,
etc.), on average it takes no less than 1,005 seconds for our (highly
parallelized) implementation of CCA to evolve the weighting function
for each class of \textsc{Reuters-21578}, 3,235 seconds for
\textsc{20Newsgroups}, and 3,075 seconds for \textsc{Ohsumed}.}

% ---------------------------------------------

\section{Would Learning the $\dd(t,d)$ Factor also Help?}
\label{sub:learntf}

\bblue{So far we have framed the LTW framework as one that learns the
IDF-like component (the $\cd(t,D,C)$ factor) alone, relying on a
predefined and fixed $\dd(t,d)$ function for handling the term
frequency component.  Despite the fact that the above is consistent
with previous STW literature, it might seem legitimate to also try to
learn the $\dd(t,d)$ component. In fact, one might argue for
explicitly optimizing also the $\dd(t,d)$ factor (instead of using
e.g., a predefined log-scaled function) by saying that the impact of
term frequency on the importance of a feature might in principle be
captured by functions different from the ones routinely used for this,
e.g., by functions that are not necessarily monotonic.  In this
section we report on experiments we have conducted in order to
ascertain whether explicitly optimizing also the $\dd(t,d)$ factor
might be beneficial.}

\bblue{To this aim, we have investigated a variant of the Local LTW
architecture (see Figure \ref{fig:LTW-arch}(a)) in which the left
branch is replaced by a component modelled by equations
\begin{equation}
  \begin{aligned}
    \hat{h}_t  & = \ReLU([f_{td}]\times W_4 + b_4) \\
    dd(t,d) & = \ReLU(\hat{h}_t \times W_5 + b_5)
    \label{eq:localtf}
  \end{aligned}
\end{equation}%
where $W_i$ and $b_i$ (for $i\in \{4,5\})$ are the new parameters to
be learned alongside the previously defined set of parameters $W_i$
and $b_i$ (for $i\in \{1,2,3\})$.  In order to preserve sparsity, we
force $dd(t,d)$ to be 0 if feature $t$ does not occur in document $d$.
% Note that, in this case, we did not add any bias to the projection.
% The reason for not doing so, is to preserve sparsity, i.e., to
% guarantee that $dd(t,d)=0$ if the feature $t$ does not occur in
% document $d$.
}

\bblue{For this experiment we consider 100 hidden units in layer
$\hat{h}$, which yield $(1\times 100 + 100) + (100\times 1 + 1)=301$
additional parameters, %with respect to LTW-L,
and apply a 0.2 dropout to neurons in the hidden layer. We leave the
rest of the parameters untouched.}

\bblue{Table \ref{res:learntf:reducedtable} shows the results we have
obtained for this variant (here denoted
LTW-TF). %both for the bounded ($\sigma$) and unbounded (I) modalities.
For the sake of clarity, we only report the results for the SVM
learner (all other learners displayed similar patterns), together with
the percentage of relative improvement with respect to the LTW
counterpart (we here choose the LTW-L-I variant) that does not
optimize the $\dd(t,d)$ component.}

\bblue{\begin{table}[t!]  \centering
  % \resizebox{\columnwidth}{!}{%
  \begin{tabular}{|l||cc|cc|}
    \hline
    & \multicolumn{2}{c|}{$F^{M}_{1}$} & \multicolumn{2}{c|}{$F^{\mu}_{1}$}\\ \hline\hline
    \textsc{Reuters-21578} & 0.586 & (-6.84\%) & 0.865 & (-1.04\%) \\\hline
    \textsc{20Newsgroups} & 0.623 & (-1.01\%) & 0.645 & (-0.70\%) \\\hline
    \textsc{Ohsumed} & 0.639 & (-1.52\%) & 0.685 & (-0.41\%) \\\hline
  \end{tabular}%
  % }
  \caption{\bblue{LTW-TF performance and relative improvement with respect to LTW-L-I.}}
  \label{res:learntf:reducedtable}
\end{table}
} The results show that explicitly learning the $\dd(t,d)$
component does not bring about any advantage. Actually, doing it
degrades performance (and, according to the Wilcoxon signed-rank test,
this difference is statistically significant at confidence level
$\alpha=0.005$). This might come as a surprise, given that LTW-TF has
more parameters than LTW-L-I, which means it could well have learned
the log-scaled version that LTW-L-I uses if this is indeed the best
choice.

\begin{figure}[tb]
  \centering
  \includegraphics[width=0.8\columnwidth]{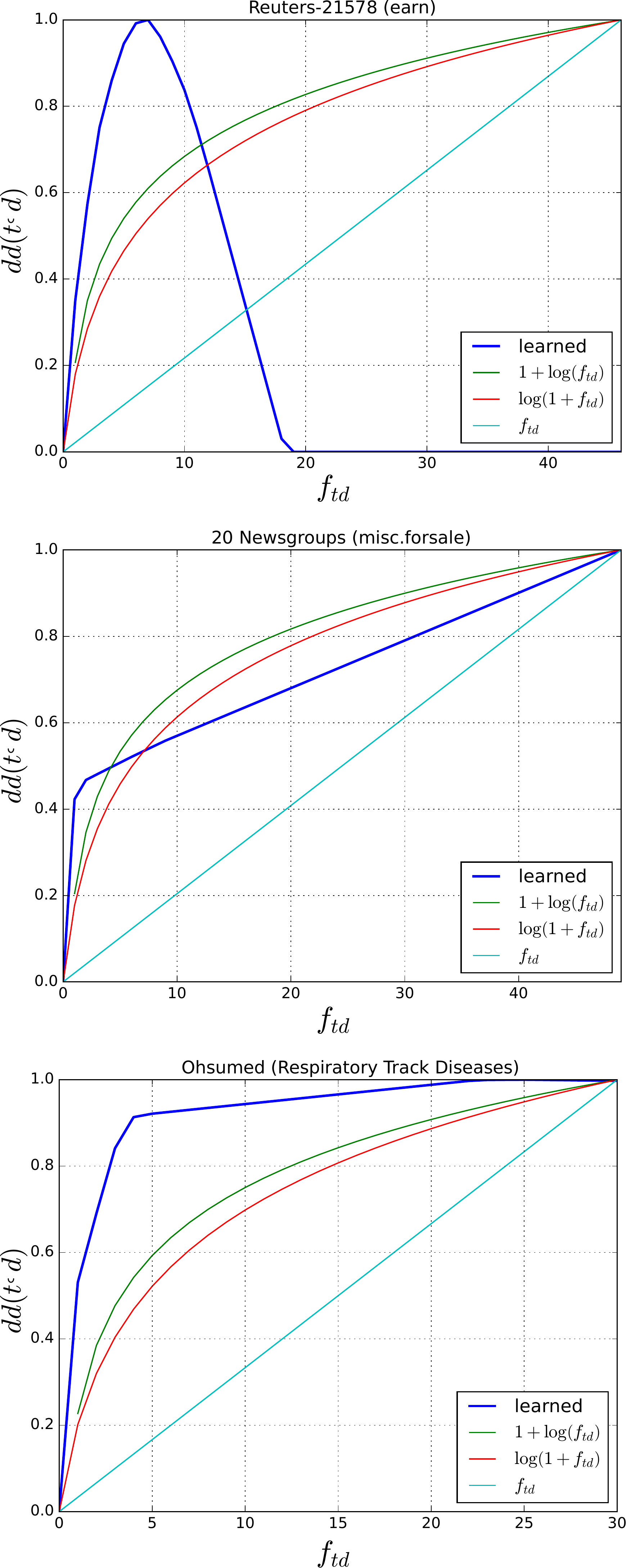}
  \caption{Learned TF-like functions for LTW-TF on (from top to
  bottom) \textsc{Reuters-21578} (earn), \textsc{20Newsgroups}
  (misc.forsale), \textsc{Ohsumed} (Respiratory Track Diseases).}
  \label{fig:tflike}
\end{figure}

So, why is LTW-TF not superior? In principle, one might think that the
proposed model is inadequate, which would imply that LTW-TF is not
able to learn any meaningful $\dd(t,d)$ function.  However, that this
is not the case can be seen by inspecting the actual functions that
the model learns, that do seem meaningful. As an example, Figure
\ref{fig:tflike} plots the functions learned for the same example
classes used for Figures \ref{fig:plotsofLTW-L} and
\ref{fig:convergence}, and compares them with typical functions used
to instantiate the $\dd(t,d)$ factor (all functions are normalized so
as to facilitate their comparison). Indeed, the functions that the
model learns seem meaningful. For example, for the
\textsc{Reuters-21578} example class (see Figure
\ref{fig:tflike}(top)) the learned function is increasing between low
($f_{td}=0$) to middle frequencies ($f_{td}=7$), then decreases until
$f_{td}>20$ and is 0 thereafter; this pattern is aligned with Luhn's
intuition that the most important words are characterized by
intermediate frequencies while words that are too rare or too frequent
should instead be attributed low importance.  In the case of
\textsc{20Newsgroups} (see Figure \ref{fig:tflike}(middle)) the
learned function displays a monotonic and quasi-linear behaviour from
approximately 0.5 onwards; this resembles another well-known
instantiation of the $\dd(t,d)$ factor, i.e.,
% the augmented and normalized term frequency (described
the $0.5+\frac{0.5+f_{td}}{max(f_{td})}$ function already documented
in past IR literature (see, e.g., \cite{Baeza-Yates:2011xd,Salton88}).
Also in \textsc{Ohsumed} (see Figure \ref{fig:tflike}(bottom)) the
model seems to have found meaningful patterns, somehow resembling the
logarithmic variants of the $\dd(t,d)$ factor.  We have observed
similar (and analogously meaningful) patterns for the other classes
too.  In sum,
% although the question about whether or not these functions are the
% best possible models for the term frequency remains unanswered,
it is clear that the degradation in performance observed in Table
\ref{res:learntf:reducedtable} cannot be explained by the supposed
inability of the $\dd(t,d)$ branch of the LTW-TF architecture to learn
meaningful functions.
\bblue{Rather, our conjecture is that the increase in the number of
parameters complicates the optimization
procedure %to an extent which is greater than the gain in model flexibility; Yet another possible explanation may be that
more than it improves the model flexibility, and that the net effect
is a model more difficult to optimize.
% This experiment thus seems %reveal
% to reinforce the belief that variations in frequency
%% of a term in a document
% bring about information which is only secondary to quantify term
% relevance, and thus that the principal information resides in the
% distribution of the term across the classes.
}

% ---------------------------------------------

\section{Conclusions}
\label{sec:conclusions}

\noindent While standard (unsupervised) term weighting approaches do
not leverage the distribution of the term across the classes,
supervised ones are able to exploit this information. However, the
improvements that these supervised methods have shown with respect to
their unsupervised variants have not been, so far, systematic.  After
discussing the possible causes of this,
% \fabscomment{Have we?},
we have discussed our conjecture that a pre-defined ``Holy Grail''
formula for supervised term weighting, after all, may not exist.
Based on this intuition we have proposed ``Learning to Weight'', a
framework for learning a supervised term weighting function tuned on
the available training set.  We have shown several instantiations of
this framework to consistently outperform previously propised
(unsupervised or supervised) term weighting approaches, on several
standard datasets and using different learning algorithms for training
classifiers.  The analysis of the weighting functions that our
algorithm has learned supports our hypothesis that the optimal
geometrical shape of the function is dependent on the underlying data
distribution.

% \strikeout{The approaches we have proposed present some limitations
% with respect to traditional term weighting approaches, e.g., the
% optimization process increases the computational cost, the learnt
% functions may fail to generalize well on small datasets, and
% hyperparameters require optimization.  Nevertheless, this first
% analysis of the ``Learning to Weight'' framework served well to
% verify our hypothesis, and will serve as the foundation for future
% work that might also attempt to lower the computational cost.}

From the point of view of the optimization processes (in particular:
the architecture of the neural networks) our approach presents some
novelty as well, since the $\cd(t,D,C)$ part of the supervised term
weighting function may be thought of as a regulariser operating on
feature statistics.  That is, the architecture we propose can be
interpreted as a traditional feedforward network (the $\dd(t,d)$ part
and logistic regressor) which is regularised (through the $\cd(t,D,C)$
part) with constant feature statistics (extracted from the matrix
columns) that control the information flow of the examples (i.e., the
matrix rows). In the future it may be worthwhile to experiment with
such kind of regulariser outside the scope of text classification.

Directions for future work include investigating the inclusion of more
elaborated statistics about the correlation between features and
classes (such as, e.g., Kullback-Leibler divergence, Fisher
Information, or other feature scoring functions).  We also plan to
test ``Learning to Weight'' in multiclass classification settings,
i.e., by
% producing a single weighted version of the dataset which takes into
% consideration
considering all classes jointly via global policies.  It would also be
interesting to extend the framework so as to jointly learn not only
the $\dd(t,d)$ function but also its normalisation
component. %, or to replace the logistic regressor part with the specific learner device , e.g., an SVM .
Finally, we believe that ``Learning to Weight'' might in principle
also be useful in other tasks related to text classification, such as
in learning to rank or feature selection. \fabscomment{Checked, Aug
14.}
% \fabscomment{Why feature selection?}

% ----------------------------------------------------------------------------

% \ifCLASSOPTIONcaptionsoff
% \newpage
% \fi

% ----------------------------------------------------------------------------
% \vspace{2cm}
% 
\bibliographystyle{IEEEtran} \bibliography{Fabrizio}

\end{document}